\documentclass[runningheads]{llncs}

\usepackage{eccv}

\usepackage{eccvabbrv}
\usepackage{multirow} 
\usepackage{graphicx}
\usepackage[accsupp]{axessibility}  
\usepackage{booktabs}
\usepackage[table]{xcolor}   
\usepackage{subcaption}

\usepackage{hyperref}

\usepackage{orcidlink}

\begin{document}

\title{Beyond Fixed Luminance: Towards Panchromatic and Orthochromatic Image Colorization} 

\titlerunning{Towards Panchromatic and Orthochromatic Image Colorization}

\author{Swarnim Maheshwari\inst{1} \and
Syed Imam Ali\inst{1} \and
Vineeth N. Balasubramanian\inst{1,2}}

\authorrunning{S.~Maheshwari et al.}

\institute{Indian Institute of Technology Hyderabad, Kandi, Telangana, India\\
\email{\{cs25mtech02006,ai24mtech14005\}@iith.ac.in}
\and
Microsoft Research India, Bengaluru, Karnataka, India\\
\email{vineethnb@cse.iith.ac.in, vineeth.nb@microsoft.com}}

\maketitle

\begin{abstract}
  Most image colorization systems operate in $Lab$ space by predicting chroma ($ab$) while preserving an input-derived luminance channel ($L$). While effective on standard benchmarks, this fixed-luminance design restricts brightness changes and becomes unreliable when grayscale formation deviates from natural-image luminance, as in historical orthochromatic photography. We propose a luminance-agnostic colorization framework that formulates colorization as full-RGB image editing using a foundation image-editing model. To bridge modern panchromatic and historical orthochromatic conditions, we introduce a mixed grayscale objective that trains the model under both standard luminance grayscale and a red-insensitive grayscale formation. Experiments on COCO, ImageNet, and a multi-instance benchmark show that our method is competitive on standard grayscale inputs and substantially more robust under orthochromatic inputs, with qualitative comparisons and a human study indicating fewer visible color artifacts.
 
\end{abstract}

\section{Introduction}
\label{sec:intro}

\begin{figure*}[t]
    \centering
    \includegraphics[width=\linewidth]{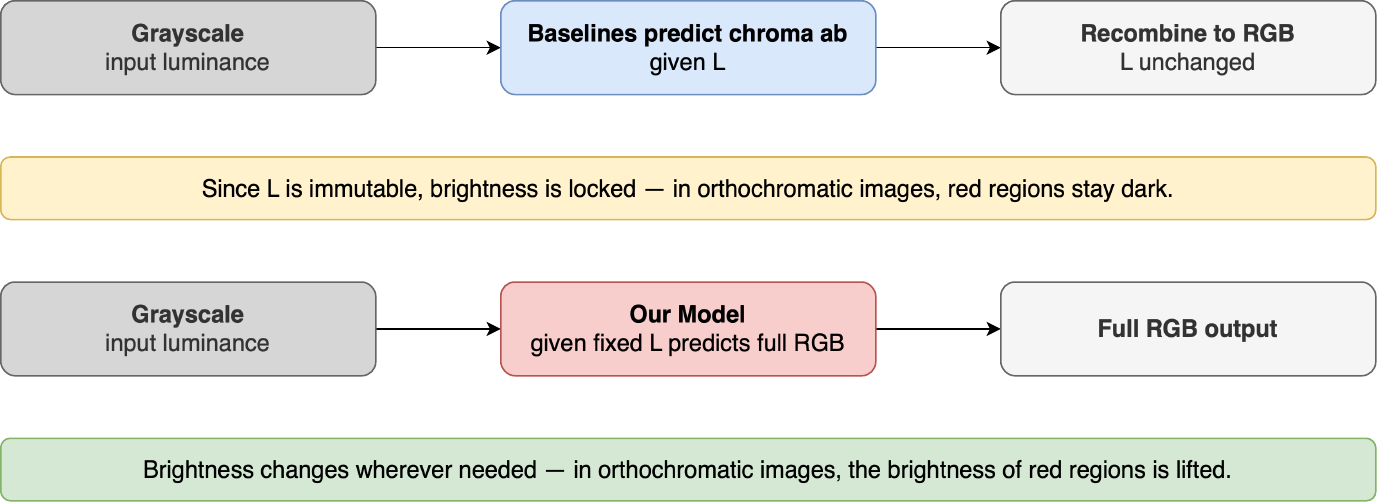}
    \caption{Comparison of the conventional colorization paradigm (\textbf{top}) with our luminance-agnostic approach (\textbf{bottom}). 
    }
    \label{fig:teaser}
\end{figure*}

The paradigm across modern approaches for image colorization follows a consistent formulation: convert an RGB image to the $Lab$ color space \cite{CIELab}, preserve the luminance channel $L$, and predict chrominance components $ab$ conditioned on that fixed luminance. This design reduces the learning problem to chroma estimation while treating luminance as ground truth. While effective for reconstruction benchmarks, this formulation imposes a strict and often overlooked constraint: the model is mathematically incapable of altering brightness (Figure~\ref{fig:teaser}). Because $L$ is treated as immutable, the network can only produce colors that are consistent with the original grayscale intensity. This restriction fundamentally limits creative recolorization. For example, mapping a dark grayscale region to a vibrant yellow or red surface is physically infeasible under fixed-$L$ constraints, since bright colors require high luminance energy in RGB space (see the d. Rose and e. Fire Ext. columns of Figure~\ref{fig:qual2}). Thus, current methods operate under a constrained prediction regime rather than generative restoration.

The issue becomes more severe when applied to historical photography. Early orthochromatic film stock was insensitive to red wavelengths, producing grayscale images whose luminance distribution does not correspond to human visual perception. Modern colorization models, trained on panchromatic data and conditioned on preserved luminance, inherit a structural domain mismatch. They are forced to assign chroma to brightness values that were never perceptually accurate, leading to suppressed reds. In this sense, current systems fail by design: when the input grayscale image is darker than it should be, the model is incapable of brightening those regions (Figure~\ref{fig:qual2}).

Most contemporary approaches that adapt pretrained text-to-image diffusion models do so by changing the architecture to condition them on grayscale \cite{controlnet,piggybacked,coco-lc,CtrlColor,bozic2024versatile, lcad}. While somewhat effective, these approaches inherit a fundamental mismatch: text-to-image models are not pretrained for image-conditioned tasks. As a result, significant architectural engineering is required to ``force'' grayscale structure into a generative pipeline originally designed for synthesis from text alone. Recent foundation models natively trained on image editing provide a superior alternative for colorization without architectural hacking~\cite{labs2025flux1kontextflowmatching,wu2025qwenimagetechnicalreport,flux-2-2025}.

In this work, we propose a luminance-agnostic framework for image colorization. Instead of predicting chroma conditioned on fixed $L$, we fine-tune a foundation image-editing model to predict the full RGB output directly from grayscale input—without luminance swapping. Our approach leverages the strong generative and editing prior of the FLUX.2-klein (4B) model. We choose FLUX.2-klein because it supports high-quality few-step generation, reducing inference time to a few seconds per image. 
Our key contributions are three-fold: (i) we identify the fixed-luminance bottleneck in Lab-based colorization and show why it is especially problematic for orthochromatic imagery; (ii) we propose a luminance-agnostic colorization framework that formulates colorization as full-RGB image editing using a foundation image-editing model; and (iii) we perform a comprehensive suite of experiments which show that the proposed method, while remaining competitive on standard grayscale inputs, is substantially more robust on orthochromatic inputs and also improves prompt-following behavior compared to baselines.

\section{Related Work}

\subsection{Image colorization}
Among early \textbf{CNN colorizers}, CIC~\cite{cic} framed colorization as classification in quantized CIELAB to avoid desaturated outputs; InstColor~\cite{instColor} colorizes instance crops with a fusion module (but can suffer from erroneous external priors and color overflow); and DISCO \cite{DISCO} uses a coarse-to-fine, anchor-based scheme to learn global affinities and reduce color ambiguity.

\textbf{GAN-based colorization} methods trade realism, speed, and cost: ChromaGAN \cite{ChromaGAN} adopts a PatchGAN discriminator with a WGAN loss for realism, HistoryNet \cite{HistoryNet} augments generation with classification/segmentation modules and a large old-movie dataset, DeOldify \cite{DeOldify} speeds training via asynchronous generator/discriminator updates, ToVivid \cite{tovivid} uses pretrained BigGAN \cite{biggan} for inversion but suffers from inversion inaccuracies on grayscale inputs, and BigColor~\cite{bigcolor} embeds BigGAN’s~\cite{biggan} generator/discriminator into an encoder–generator model.
Time-Travel Rephotography~\cite{Luo-Rephotography-2021} projects old portrait photographs into the manifold of modern high-resolution face images using a StyleGAN2 prior, jointly performing restoration, enhancement, and recolorization; notably, it also considers different historical film types, including orthochromatic negatives, although it is specialized to face portrait imagery.

\textbf{Transformer-based colorization} has advanced quickly: ColTran \cite{coltran} proposed a multi-stage transformer colorizer but suffers from limited CNN-like inductive bias; ColorFormer \cite{colorformer} adds a global–local hybrid self-attention and a color memory for efficient semantic–color mapping; ct2 \cite{ct2} derives 313 meaningful color tokens from color-space statistics and uses adaptive attention to link them to luminance; DDColor \cite{ddcolor} introduces learnable color tokens with cross-attention fusion of grayscale and color features; and MultiColor \cite{MultiColor} builds on DDColor with a multi-branch design to better capture colors using different color spaces.

\textbf{Diffusion-based colorization from text-to-image backbones.}
Recent work has adapted pretrained text-to-image diffusion models for colorization by injecting grayscale structure into the generative process. Existing strategies include ControlNet-style conditioning branches \cite{controlnet,piggybacked,coco-lc,CtrlColor}, mid-layer feature injection through extended convolutional modules \cite{lcad}, and formulations that replace the initial noise with grayscale image structure \cite{Diffusing-Colors}. These methods demonstrate that large diffusion priors can improve semantic colorization quality, but they typically require additional architectural modifications to make text-to-image models usable for image-conditioned restoration tasks.

\subsection{Image context diffusion models}

More recently, a new class of foundation generative models has been developed specifically for image editing and in-context image generation. Examples include Qwen-Image \cite{wu2025qwenimagetechnicalreport}, FLUX.1 Kontext \cite{labs2025flux1kontextflowmatching}, and FLUX.2-klein \cite{flux-2-2025}. Such models are trained to preserve and transform image content directly, making them a more natural backbone for colorization than text-to-image models that must be retrofitted with grayscale conditioning. Our method follows this direction by formulating colorization as an image-editing problem rather than as chroma prediction under a fixed luminance constraint.

\section{Methodology}

\textbf{Our overall objective}: A core contribution of our work is the Mixed Grayscale Objective, designed to bridge the gap between modern digital intensity and historical film sensitivity \cite{zerman2019colornet}.

To this end, given a ground-truth colorful image $I \in \mathbb{R}^{H \times W \times 3}$ with channels $I_R, I_G, I_B$, we generate two distinct types of grayscale conditioning images, denoted as $I_{gray}$. 

For the \textbf{Standard Mode}, we simulate modern panchromatic film using the standard luminance formulation, conditioned with the text prompt $c_{pan} = \text{``colorize'' + Caption}$:
\begin{equation}
    I_{gray}^{pan} = 0.299 \cdot I_R + 0.587 \cdot I_G + 0.114 \cdot I_B
\end{equation}

For the \textbf{Orthochromatic Mode}, we simulate historical red-insensitive film using a simplified approximation motivated by the fact that orthochromatic materials are sensitive primarily to blue and green wavelengths and insensitive to red~\cite{britannica_orthochromatic_film,filmcolors_orthochromatic_stock_1873,kodak_filmessentials_characteristics_06,kodak_mpt_glossary_ortho}. To simulate orthochromatic images from modern-day color images, we take inspiration from prior work on historical photo restoration that explicitly models different negative film types~\cite{Luo-Rephotography-2021}. We therefore exclude the red channel and define 
\begin{equation}
I_{gray}^{ortho} = \frac{I_B + I_G}{2},
\end{equation}
conditioned with the text prompt $c_{ortho} = \text{``colorize ortho'' + Caption}$. Here, ``colorize'' acts as a trigger word for the action of colorization, while ``ortho'' acts as a trigger word to notify the model that the input is an orthochromatic image.  During training, we encode the ground-truth target image into the latent space using the frozen VAE, yielding the data latent $z_1 = \mathcal{E}(I)$. Because the Flux architecture is based on Rectified Flow (a formulation of continuous-time diffusion models), the forward process constructs noisy latents $z_t$ by linearly interpolating between standard Gaussian noise $z_0 \sim \mathcal{N}(0, I)$ and the target latent $z_1$ over timestep $t \in [0, 1]$:
\begin{equation}
    z_t = t z_1 + (1 - t) z_0
\end{equation}

The DiT is trained to predict the vector field (velocity) $v_\theta$ that transports the noise distribution to the data distribution. The network is conditioned on the text embedding $c_{text} \in \{c_{pan}, c_{ortho}\}$ and the spatially concatenated grayscale image $I_{gray} \in \{I_{gray}^{pan}, I_{gray}^{ortho}\}$. The generative diffusion objective $\mathcal{L}$ is minimized as follows:
\begin{equation}
    \mathcal{L} = \mathbb{E}_{z_0, z_1, t} \left[ \| v_\theta(z_t, t, c_{text}, I_{gray}) - (z_1 - z_0) \|_2^2 \right]
\end{equation}

By training on this objective, the model learns to interpret luminance values differently based on the provided text prompt. Instead of replacing the predicted luminance with a constrained $L$ channel during inference, the network maps the conditional inputs directly to a full RGB representation. This allows it to ``lift'' the brightness of dark regions in orthochromatic inputs where a standard $L$-constrained model would otherwise fail.

\textbf{Architecture and Training Configuration}: We employ FLUX.2-klein-4B as our foundational generative prior, leveraging its robust semantic understanding for high-fidelity color synthesis. To maintain computational efficiency while adapting the model to the colorization task, we utilize Low-Rank Adaptation (LoRA) applied exclusively to the Diffusion Transformer (DiT) blocks. Both the VAE (Image Encoder/Decoder) and the Qwen3-4B Text Encoder remain frozen throughout the process. Our LoRA configuration utilizes a rank of 16 and an alpha of 16, resulting in approximately 23.10 million trainable parameters, representing a mere 0.5\% of the total 4.3B parameter count. 

\textbf{Optimization and Training Procedure}: The model was trained using the AdamW optimizer with a constant learning rate of $1 \times 10^{-4}$ and a 100-step warmup period. We utilize bfloat16 mixed-precision and enable gradient checkpointing. Training was executed across a distributed setup of three NVIDIA RTX 6000 Ada GPUs with an effective batch size of 12. The model was trained for 10 epochs on a subset of ImageNet from Kaggle \cite{kaggle_imagenet}, totaling 34,745 images.

\section{Experiments}
\subsection{Datasets}
To evaluate our proposed method across different visual domains, we utilize subsets of three widely adopted benchmarks:
    \textbf{ImageNet} \cite{imagenet}: We sample the first 5,000 images from the validation set.
    \textbf{COCO} \cite{coco-stuff}: We use the first 5,000 images from the test set.
     \textbf{Multi-Instance} \cite{lcoins}: The complete validation set, consisting of 7,213 images.
For ImageNet and COCO, we used BLIP2~\cite{li2023blip2} to generate captions, while for Multi-Instance, human-annotated captions were provided with the dataset.

\subsection{Evaluation Metrics}
We evaluate on both perceptual realism and the statistical distribution of the generated colors.

\textbf{Perceptual Quality Metrics:} We utilize the PyTorch Image Quality Assessment \cite{pyiqa} framework to compute the Fréchet Inception Distance (FID) \cite{FID}, spatial FID (sFID) \cite{sfidding2020continuous}, and FID-DINO, which is the same as FID but uses DINOv2~\cite{oquab2023dinov2} features instead of InceptionNet. These metrics measure the distributional shift between the synthesized and ground-truth images.

\textbf{Chroma and Colorfulness Metrics:} To evaluate both the perceptual vibrancy and the statistical distribution of the generated colors independent of luminance, we define a unified set of metrics. 

First, we quantify human-perceived vibrancy using the standard Colorfulness metric~\cite{hasler2003measuring} widely used in image colorization work, which is computed using the mean ($\mu$) and standard deviation ($\sigma$) of the opponent color spaces $rg = R - G$ and $yb = \frac{1}{2}(R + G) - B$:
\begin{equation}
    C = \sqrt{\sigma_{rg}^2 + \sigma_{yb}^2} + 0.3 \sqrt{\mu_{rg}^2 + \mu_{yb}^2}
\end{equation}

For a dataset of $N$ images, let $\mu_a, \mu_b$ and $\sigma_a, \sigma_b$ represent the mean and standard deviation of the $a$ and $b$ color channels in the LAB color space for a given image $I_k$:

\begin{itemize}
    \item \textbf{Col-diverse (Chroma Diversity):} Measures the amount of chromatic variance generated within an image, calculated as the Euclidean combination of the channel standard deviations:
    \begin{equation}
        \text{Col-diverse} = \frac{1}{N} \sum_{k=1}^{N} \sqrt{\sigma_{a,k}^2 + \sigma_{b,k}^2}
    \end{equation}
    
    \item \textbf{Saturation (Mean LAB Chroma):} Calculates the average per-pixel chroma in the LAB space. Let $a_p$ and $b_p$ be the channel values for pixel $p$, and $|I_k|$ be the total number of pixels in image $I_k$:
    \begin{equation}
        \text{Saturation} = \frac{1}{N} \sum_{k=1}^{N} \left( \frac{1}{|I_k|} \sum_{p \in I_k} \sqrt{a_p^2 + b_p^2} \right)
    \end{equation}
\end{itemize}

\textbf{Colornet (Perceptual Colorfulness) \cite{zerman2019colornet}:} While traditional metrics rely on heuristic formulations in opponent color spaces, we also evaluate the perceptual vibrancy of our generated outputs using ColorNet. Developed by Zerman et al., ColorNet is a Convolutional Neural Network (CNN) specifically trained on subjective human datasets to predict Mean Opinion Scores (MOS) of image colorfulness. By leveraging deep feature representations rather than simple statistical variances, this learned metric provides an assessment of chromatic intensity that aligns with the human visual system.

\textbf{Baselines}: We use the following baselines: BigColor~\cite{bigcolor}, COCO-LC~\cite{coco-lc}, DDColor~\cite{ddcolor}, DISCO~\cite{DISCO}, and UniColor~\cite{unicolor}. We use BLIP2 to generate captions for both COCO-LC and our method on the ImageNet and COCO datasets, and use the provided human-labeled captions for the Multi-Instance dataset. We evaluate all baselines by converting each image to grayscale and feeding it into the model. For the ortho setting we used $I_{gray}^{ortho}$, while for the pan setting we used the standard grayscale formulation $I_{gray}^{pan}$.

\begin{figure*}[htbp]
    \centering
    \setlength{\tabcolsep}{1pt}
    \renewcommand{\arraystretch}{0.5} 
    
    \begin{tabular}{c | m{0.22\linewidth} | m{0.22\linewidth} | m{0.22\linewidth} | m{0.22\linewidth} |}
        
        & \multicolumn{2}{c|}{\textbf{a. Car}} & \multicolumn{2}{c|}{\textbf{b. Baby}} \\
        \cline{2-5}
        & \multicolumn{1}{c|}{\small \textbf{Ortho}} & \multicolumn{1}{c|}{\small \textbf{Pan}} & \multicolumn{1}{c|}{\small \textbf{Ortho}} & \multicolumn{1}{c|}{\small \textbf{Pan}} \\
        \cline{2-5}
        
        \scriptsize \textbf{Input} &
        \includegraphics[width=\linewidth, height=0.82\linewidth]{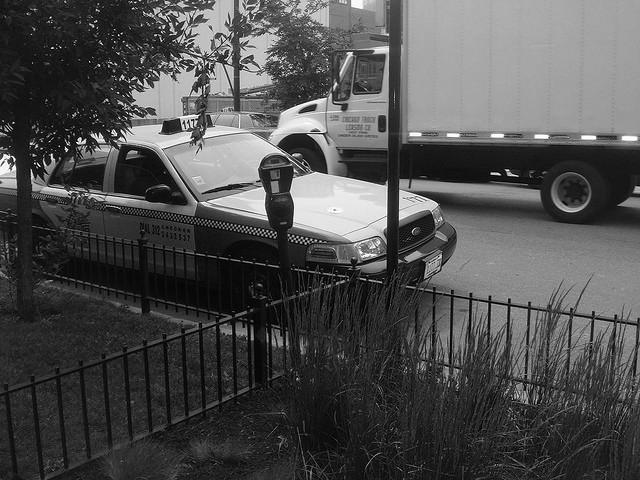} &
        \includegraphics[width=\linewidth, height=0.82\linewidth]{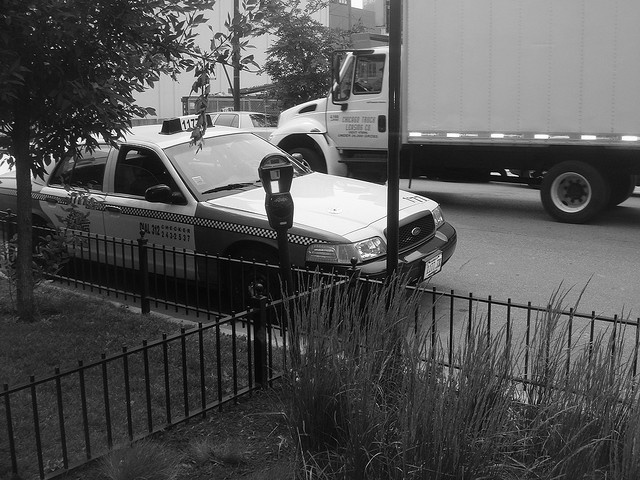} &
        \includegraphics[width=\linewidth, height=0.82\linewidth]{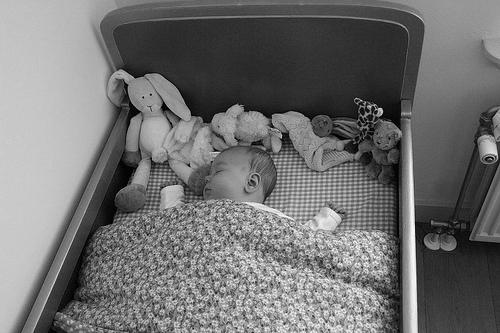} &
        \includegraphics[width=\linewidth, height=0.82\linewidth]{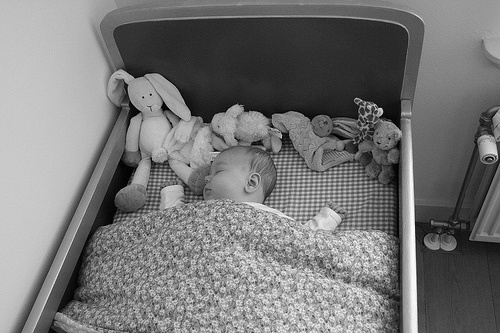} \\

        \scriptsize \textbf{DDColor} &
        \includegraphics[width=\linewidth, height=0.82\linewidth]{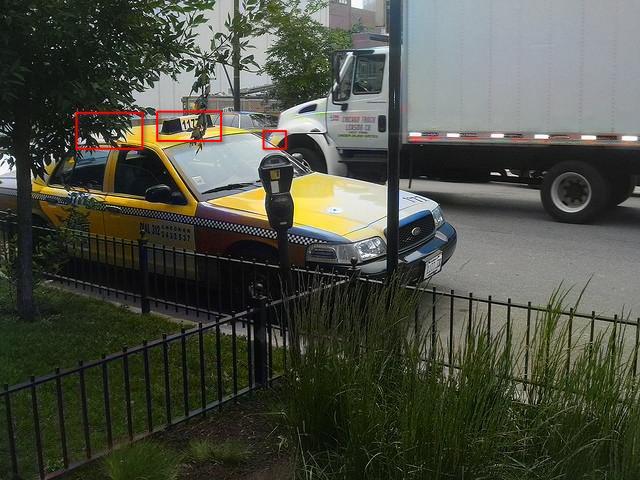} &
        \includegraphics[width=\linewidth, height=0.82\linewidth]{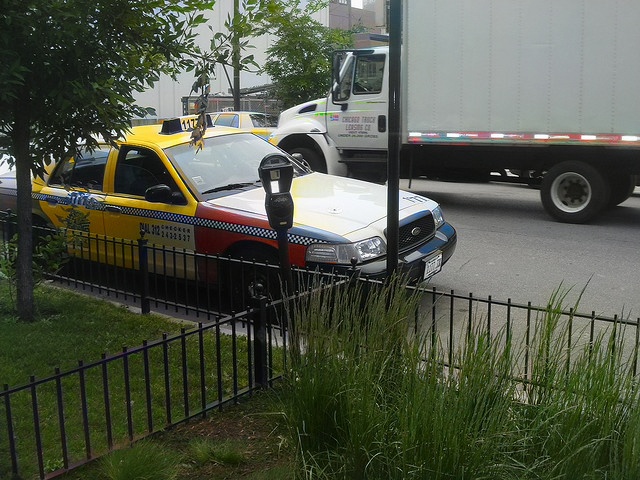} &
        \includegraphics[width=\linewidth, height=0.82\linewidth]{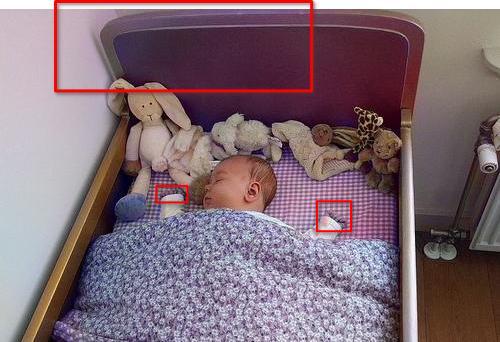} &
        \includegraphics[width=\linewidth, height=0.82\linewidth]{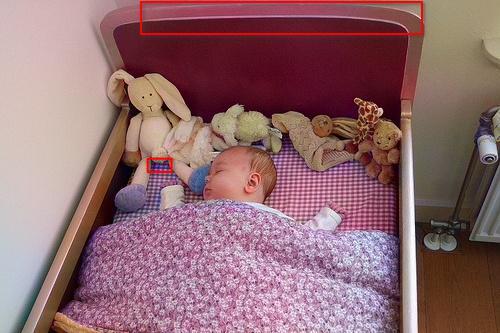} \\

        \scriptsize \textbf{DISCO} &
        \includegraphics[width=\linewidth, height=0.82\linewidth]{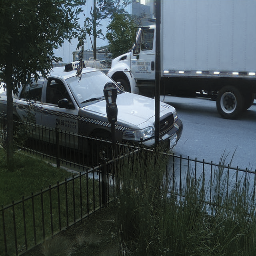} &
        \includegraphics[width=\linewidth, height=0.82\linewidth]{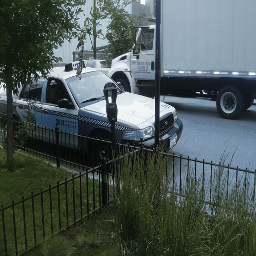} &
        \includegraphics[width=\linewidth, height=0.82\linewidth]{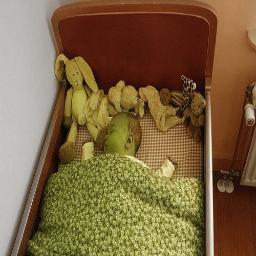} &
        \includegraphics[width=\linewidth, height=0.82\linewidth]{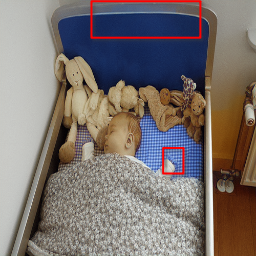} \\

        \scriptsize \textbf{COCO-LC} &
        \includegraphics[width=\linewidth, height=0.82\linewidth]{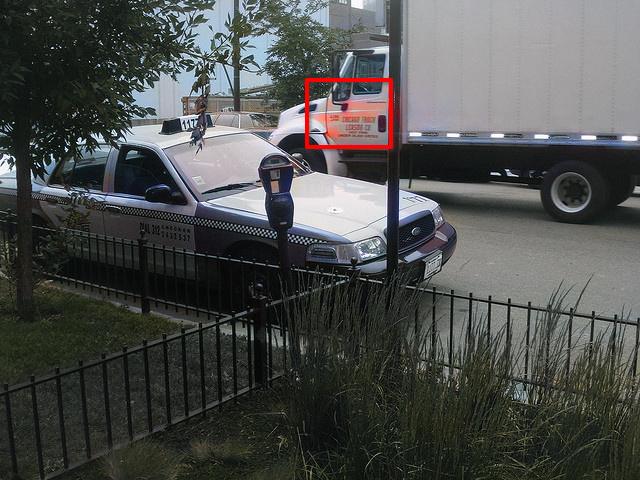} &
        \includegraphics[width=\linewidth, height=0.82\linewidth]{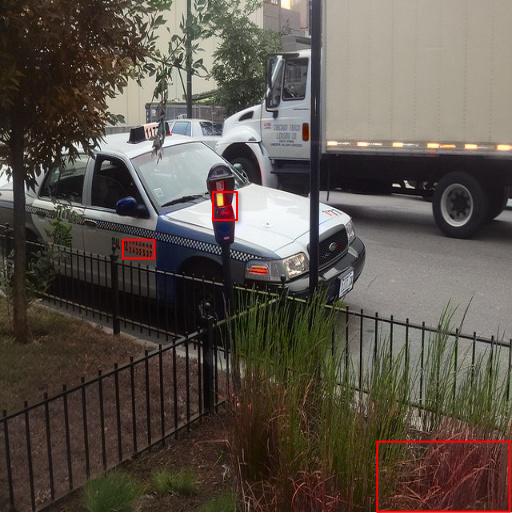} &
        \includegraphics[width=\linewidth, height=0.82\linewidth]{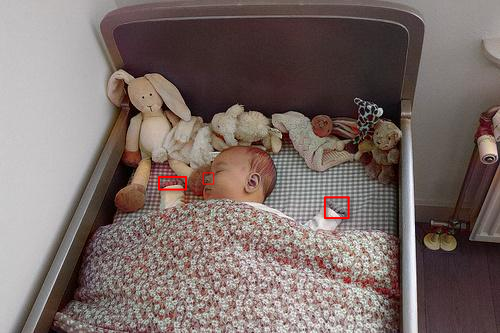} &
        \includegraphics[width=\linewidth, height=0.82\linewidth]{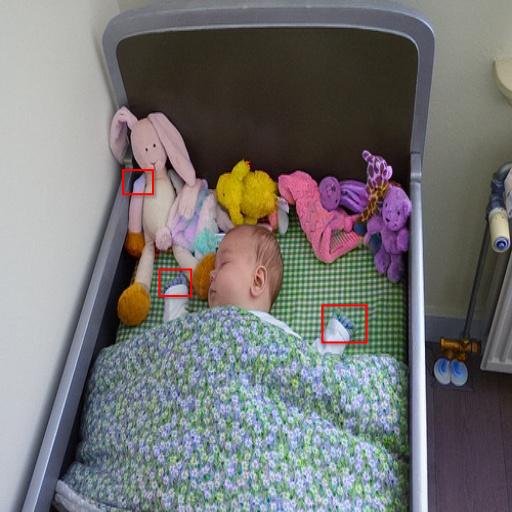} \\

        \scriptsize \textbf{BigColor} &
        \includegraphics[width=\linewidth, height=0.82\linewidth]{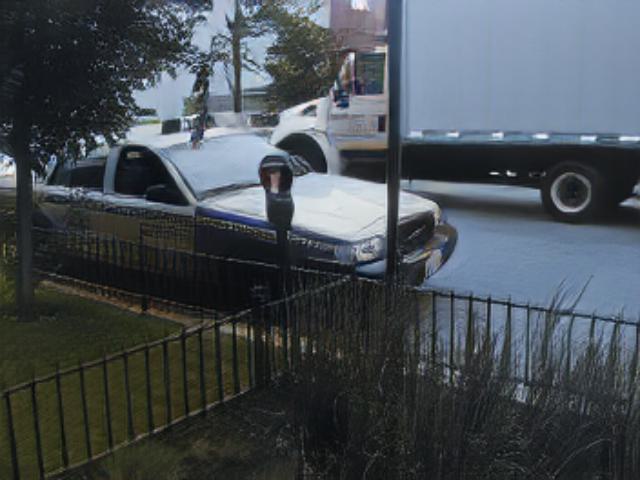} &
        \includegraphics[width=\linewidth, height=0.82\linewidth]{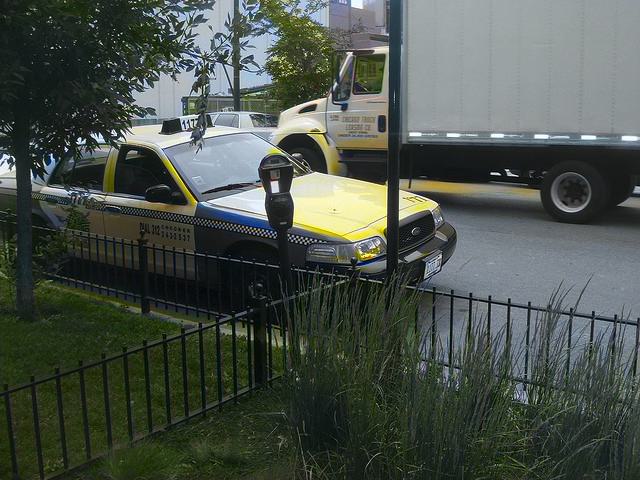} &
        \includegraphics[width=\linewidth, height=0.82\linewidth]{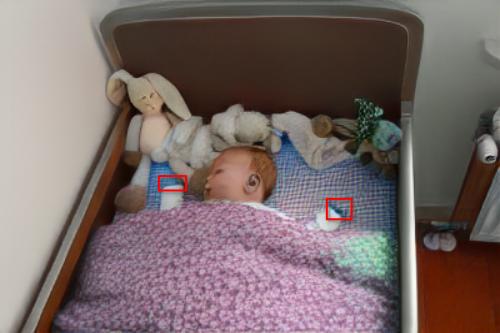} &
        \includegraphics[width=\linewidth, height=0.82\linewidth]{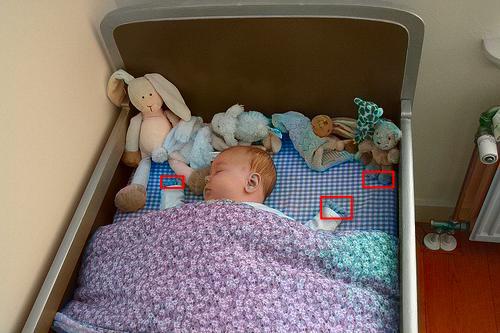} \\

        \scriptsize \textbf{UniColor} &
        \includegraphics[width=\linewidth, height=0.82\linewidth]{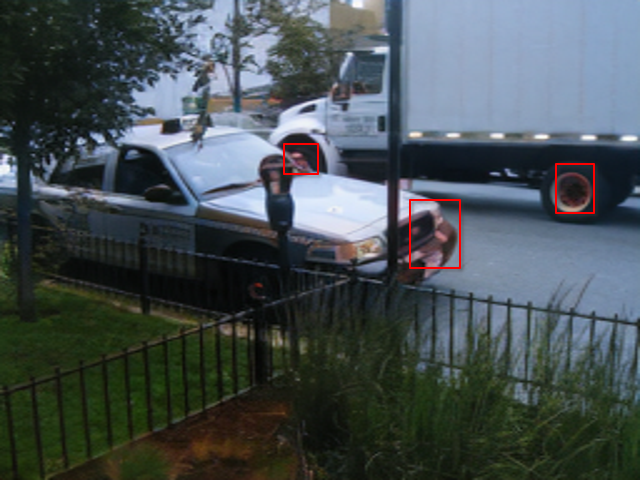} &
        \includegraphics[width=\linewidth, height=0.82\linewidth]{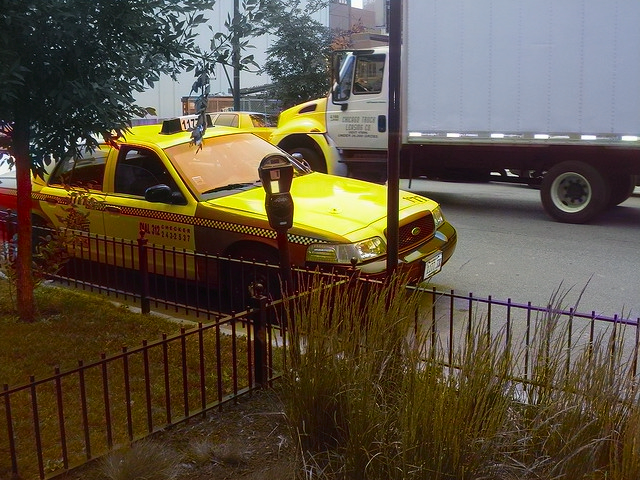} &
        \includegraphics[width=\linewidth, height=0.82\linewidth]{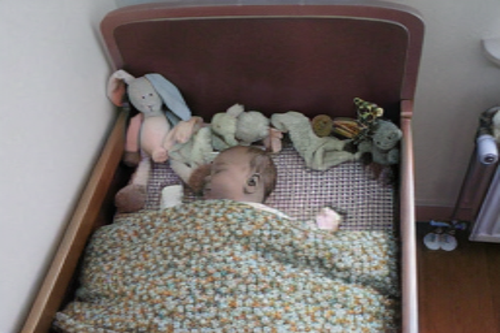} &
        \includegraphics[width=\linewidth, height=0.82\linewidth]{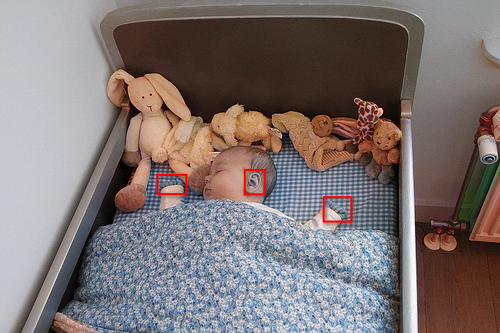} \\

        \scriptsize \textbf{Ours} &
        \includegraphics[width=\linewidth, height=0.82\linewidth]{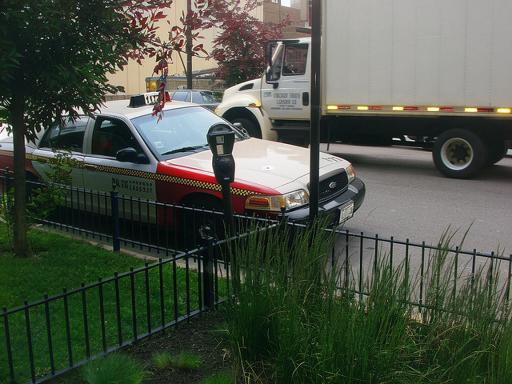} &
        \includegraphics[width=\linewidth, height=0.82\linewidth]{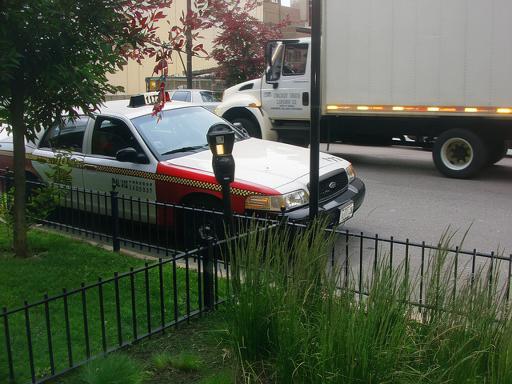} &
        \includegraphics[width=\linewidth, height=0.82\linewidth]{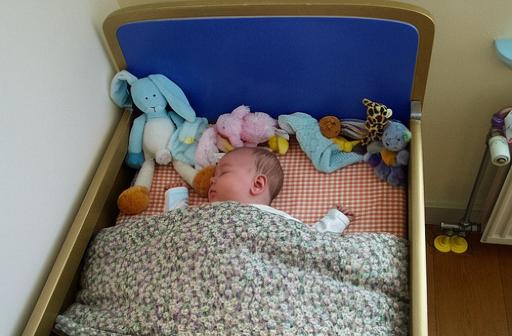} &
        \includegraphics[width=\linewidth, height=0.82\linewidth]{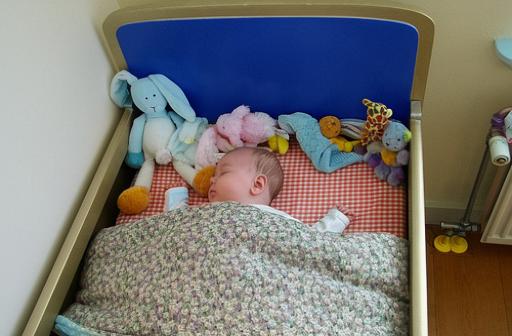} \\
    \end{tabular}
    \caption{Notice in the b. Baby images that for most methods (except ours), the color and pattern from the bedsheets bleed into the hands. \textbf{(Zoom-in for best view)}}
    \label{fig:qual1}
\end{figure*}

\begin{figure*}[htbp]
    \centering
    \makebox[\textwidth][c]{
        \setlength{\tabcolsep}{1.2pt} 
        \renewcommand{\arraystretch}{0.5} 
        
        \begin{tabular}{c | m{0.21\linewidth} | m{0.21\linewidth} | m{0.21\linewidth} | m{0.21\linewidth} |}
            
            & \multicolumn{2}{c|}{\textbf{c. Cow}} & \multicolumn{1}{c|}{\textbf{d. Rose}} & \multicolumn{1}{c|}{\textbf{e. Fire Ext.}} \\
            \cline{2-5}
            & \multicolumn{1}{c|}{\small \textbf{Ortho}} & \multicolumn{1}{c|}{\small \textbf{Pan}} & \multicolumn{1}{c|}{\small \textbf{Ortho}} & \multicolumn{1}{c|}{\small \textbf{Ortho}} \\
            \cline{2-5}
            
            \scriptsize \textbf{Input} & 
            \includegraphics[width=\linewidth, height=0.81\linewidth]{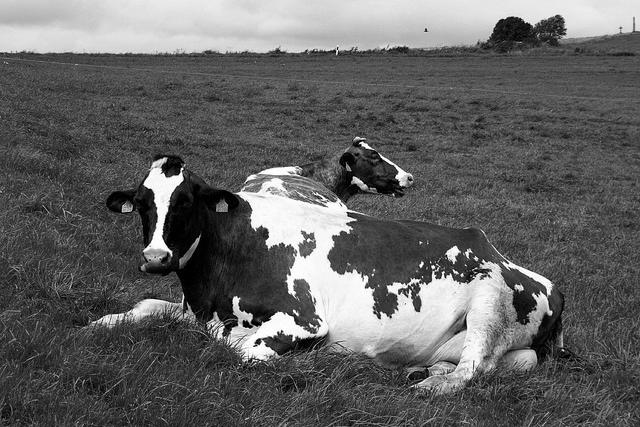} & 
            \includegraphics[width=\linewidth, height=0.81\linewidth]{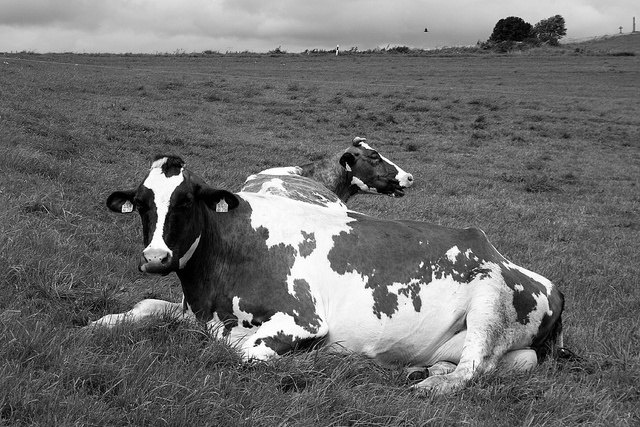} &
            \includegraphics[width=\linewidth, height=0.81\linewidth]{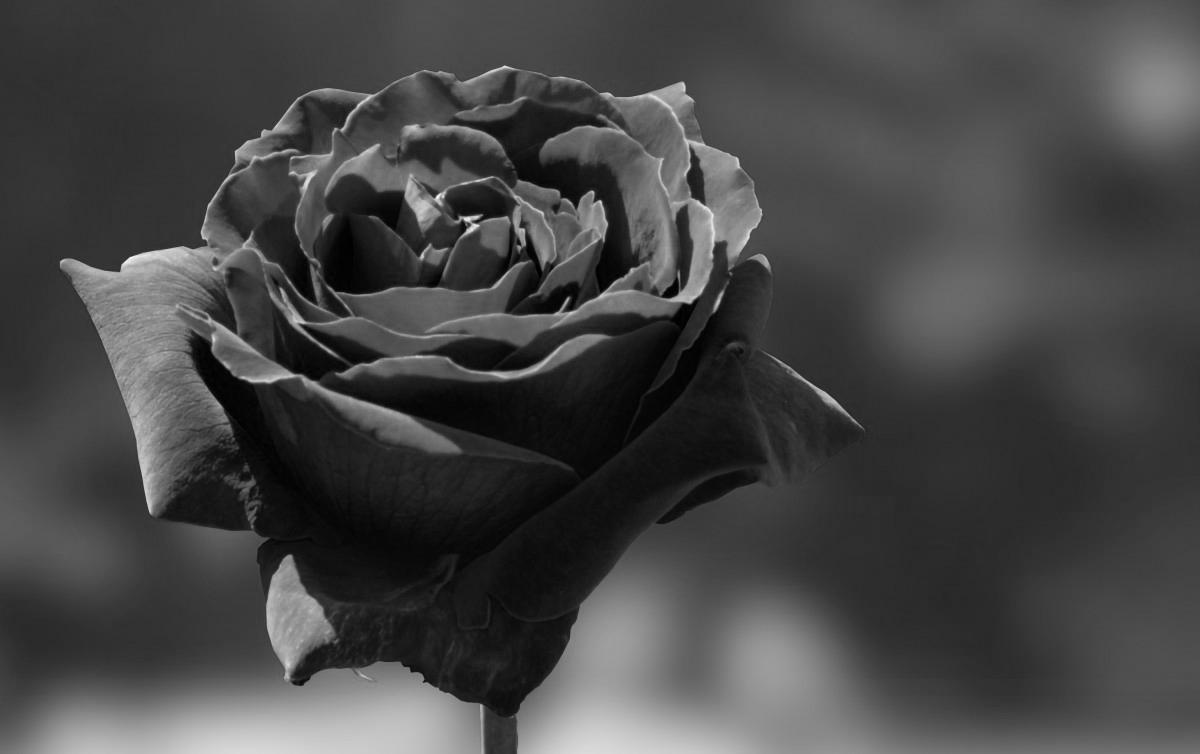} & 
            \includegraphics[width=\linewidth, height=0.81\linewidth]{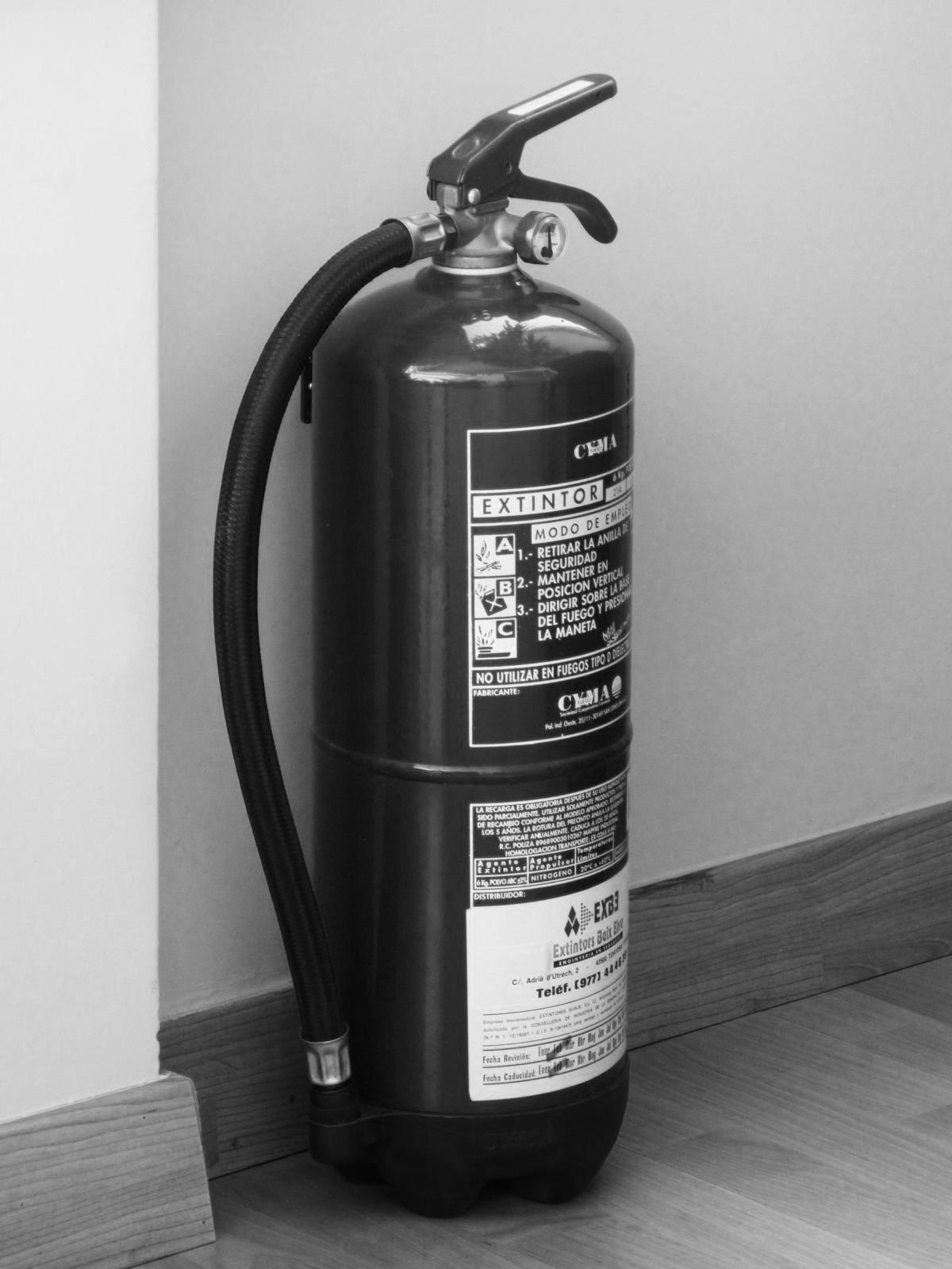} \\
    
            \scriptsize \textbf{DDColor} & 
            \includegraphics[width=\linewidth, height=0.81\linewidth]{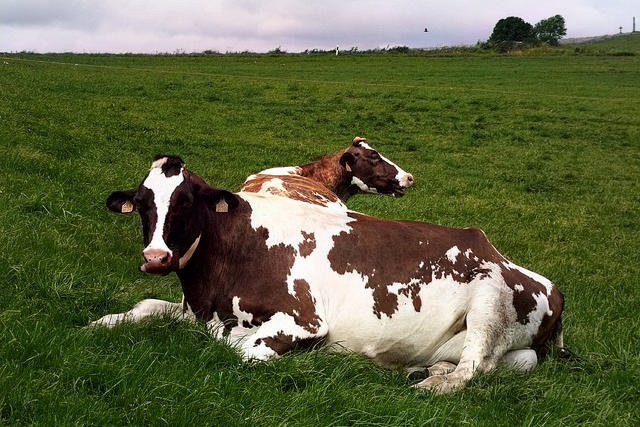} & 
            \includegraphics[width=\linewidth, height=0.81\linewidth]{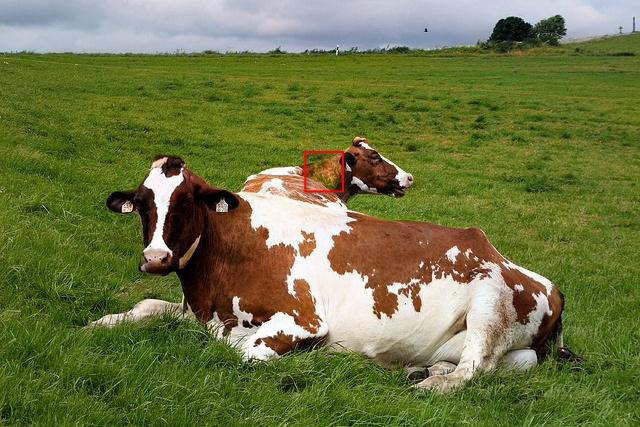} &
            \includegraphics[width=\linewidth, height=0.81\linewidth]{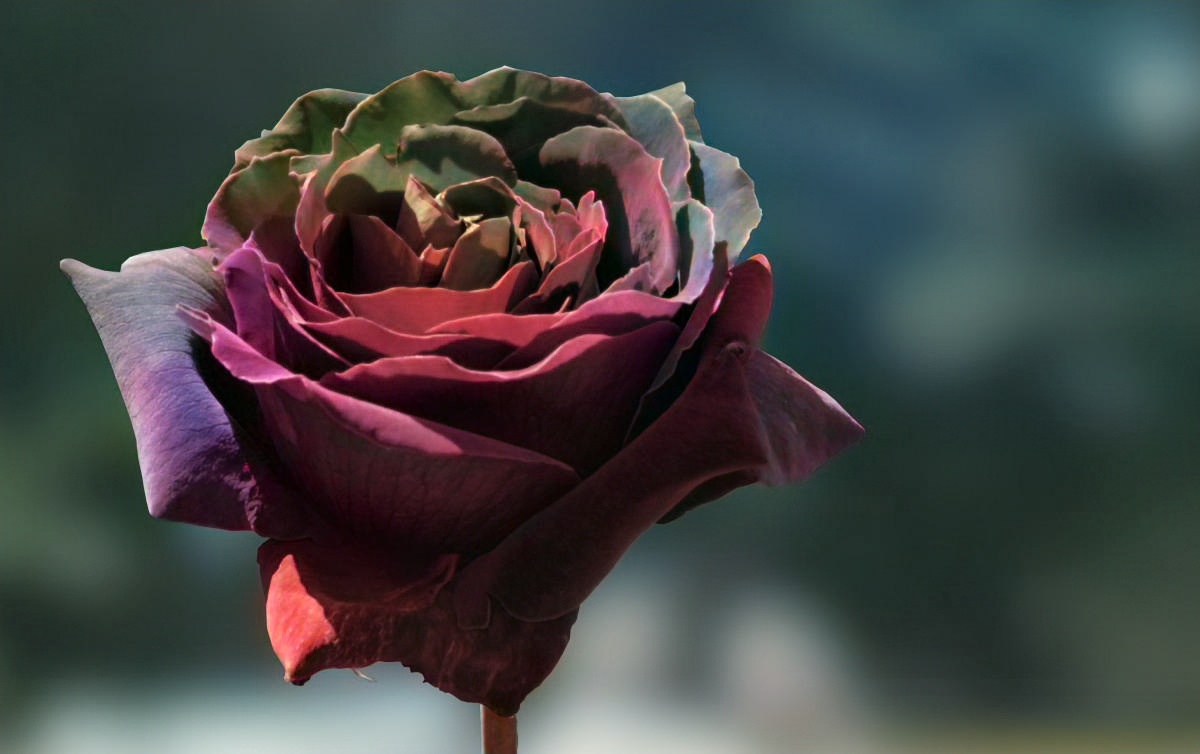} & 
            \includegraphics[width=\linewidth, height=0.81\linewidth]{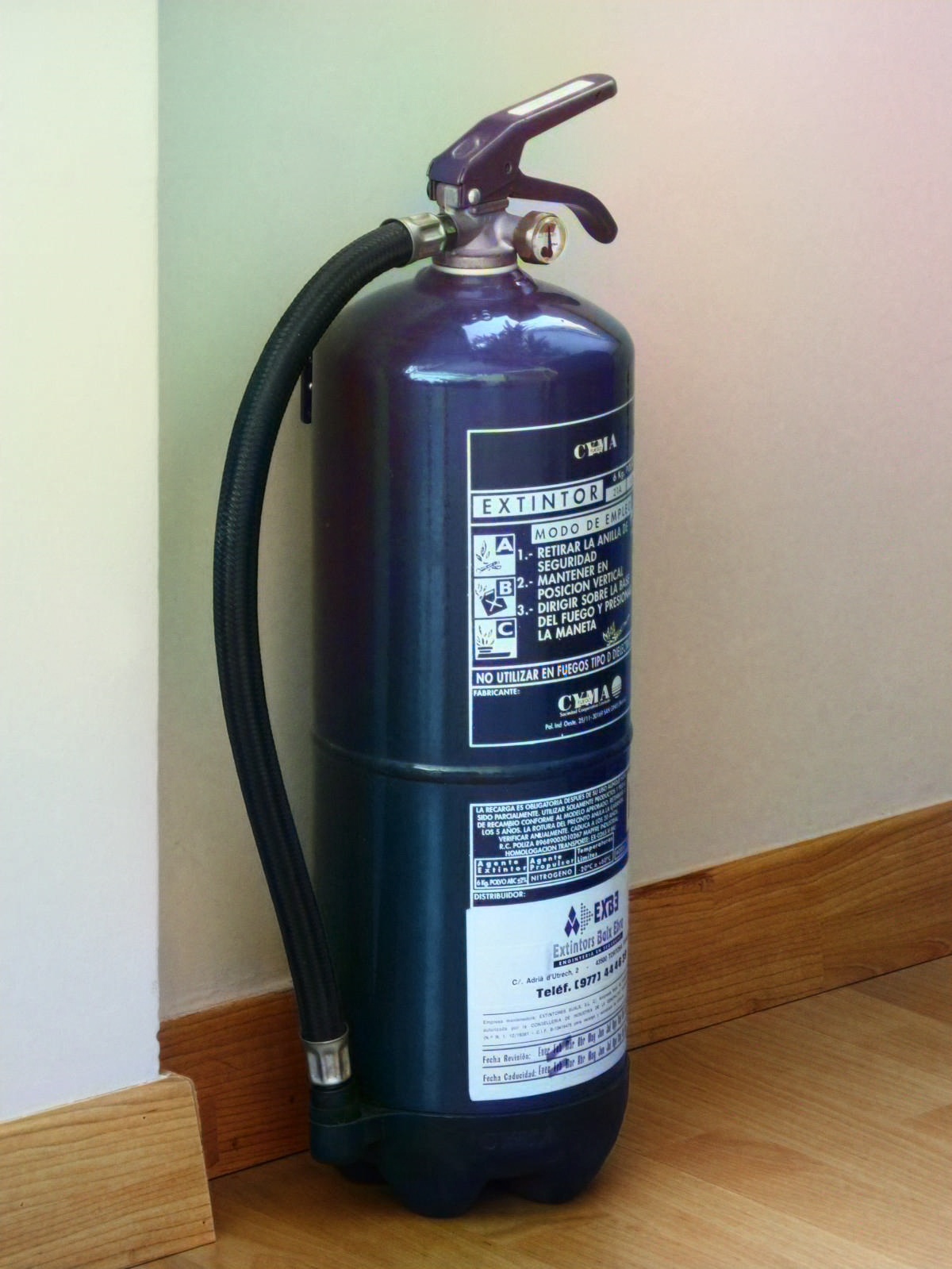} \\
    
            \scriptsize \textbf{DISCO} & 
            \includegraphics[width=\linewidth, height=0.81\linewidth]{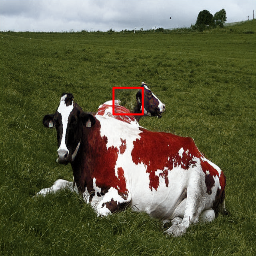} & 
            \includegraphics[width=\linewidth, height=0.81\linewidth]{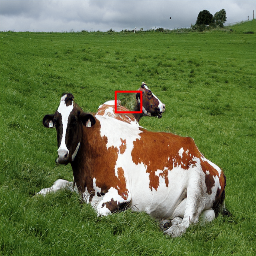} &
            \includegraphics[width=\linewidth, height=0.81\linewidth]{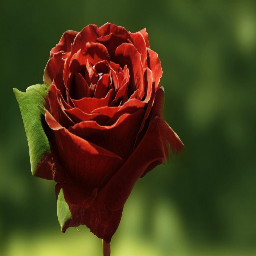} & 
            \includegraphics[width=\linewidth, height=0.81\linewidth]{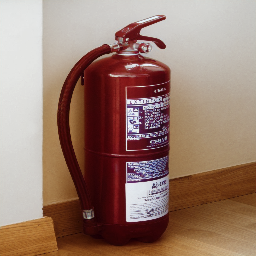} \\
    
            \scriptsize \textbf{COCO-LC} & 
            \includegraphics[width=\linewidth, height=0.81\linewidth]{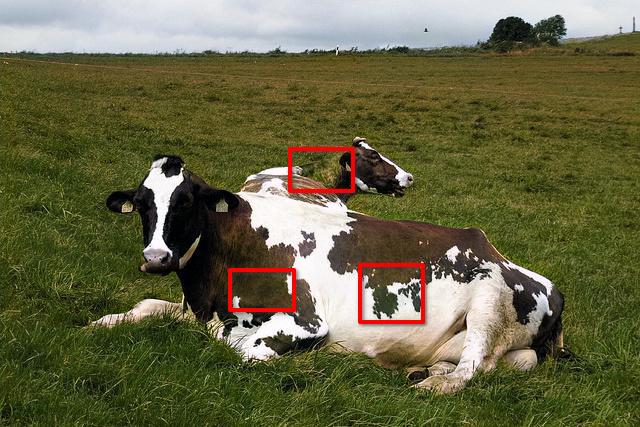} & 
            \includegraphics[width=\linewidth, height=0.81\linewidth]{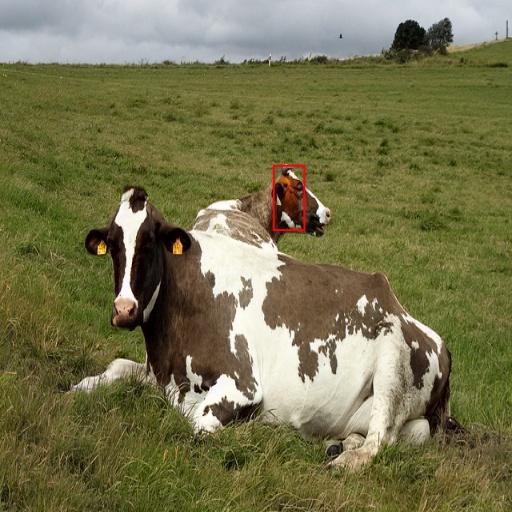} &
\includegraphics[width=\linewidth, height=0.81\linewidth]{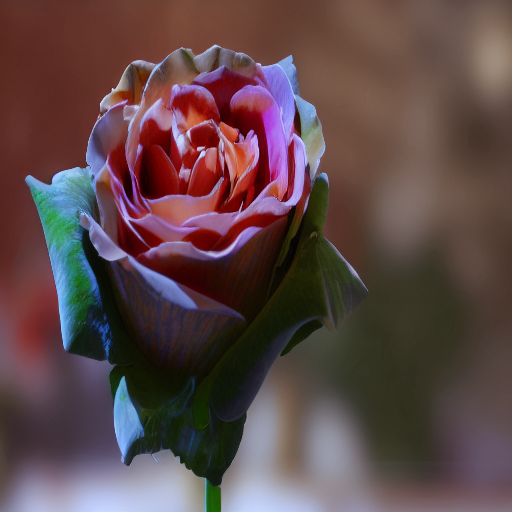} &
            \includegraphics[width=\linewidth, height=0.81\linewidth]{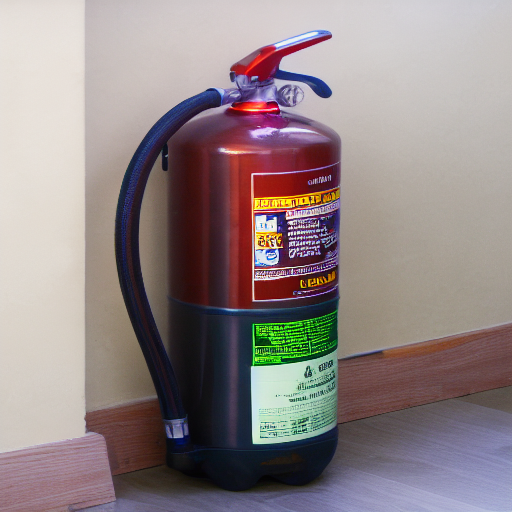} \\
    
            \scriptsize \textbf{BigColor} & 
            \includegraphics[width=\linewidth, height=0.81\linewidth]{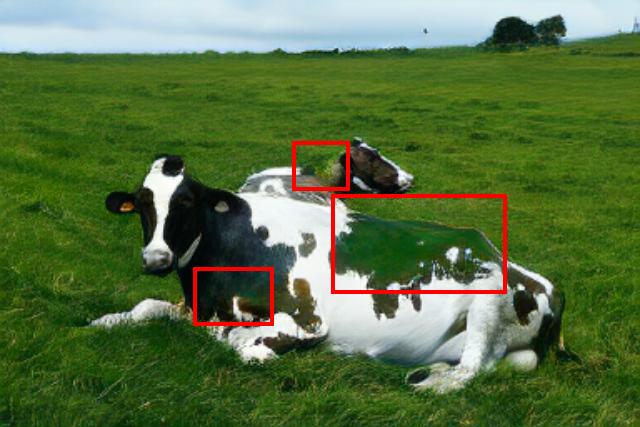} & 
            \includegraphics[width=\linewidth, height=0.81\linewidth]{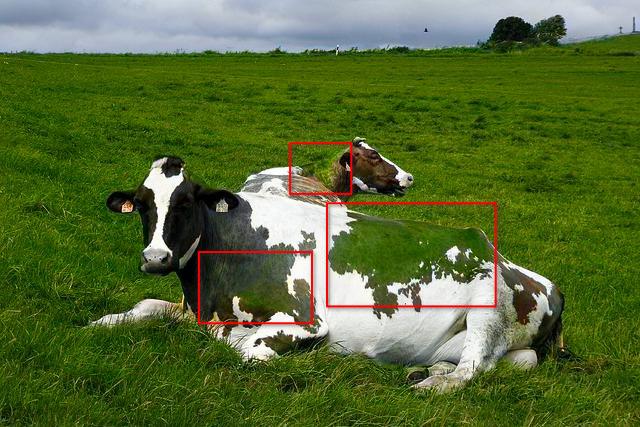} &
            \includegraphics[width=\linewidth, height=0.81\linewidth]{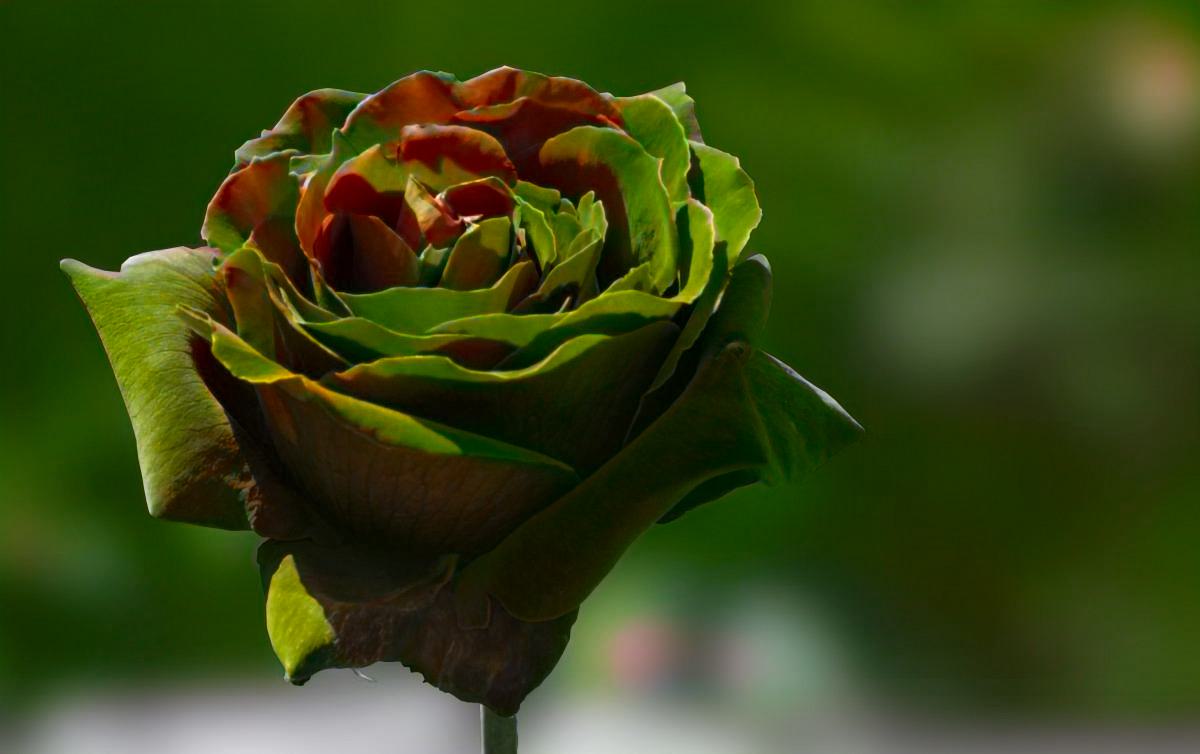} & 
            \includegraphics[width=\linewidth, height=0.81\linewidth]{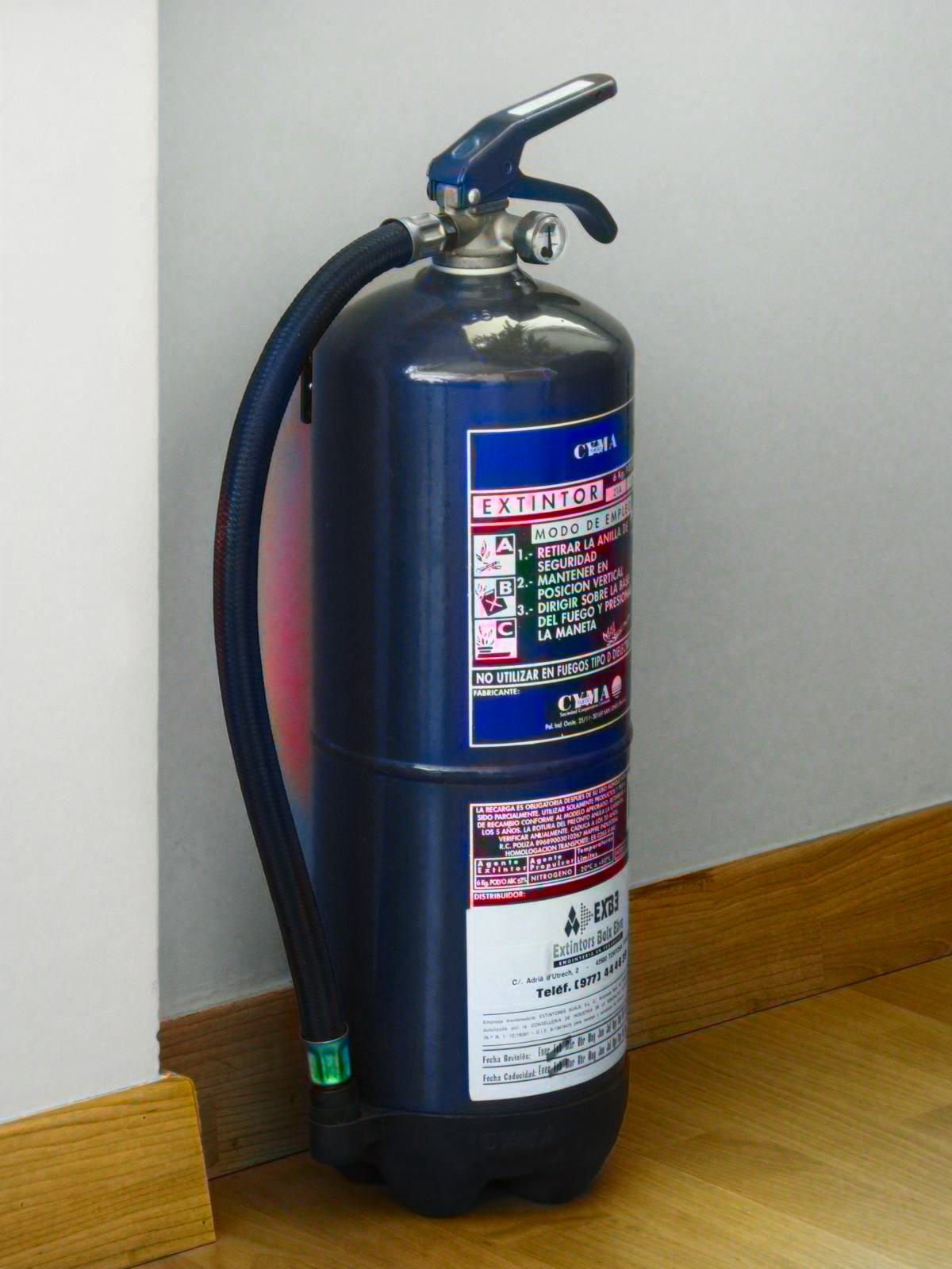} \\
    
            \scriptsize \textbf{UniColor} & 
            \includegraphics[width=\linewidth, height=0.81\linewidth]{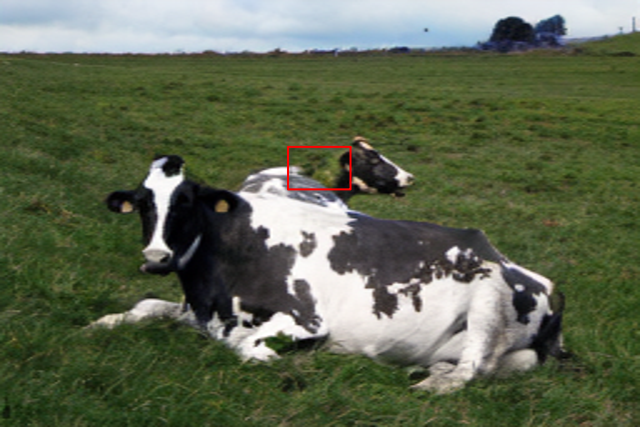} & 
            \includegraphics[width=\linewidth, height=0.81\linewidth]{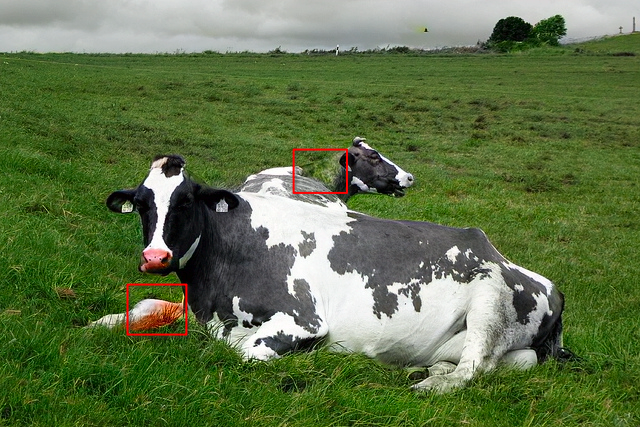} &
            \includegraphics[width=\linewidth, height=0.81\linewidth]{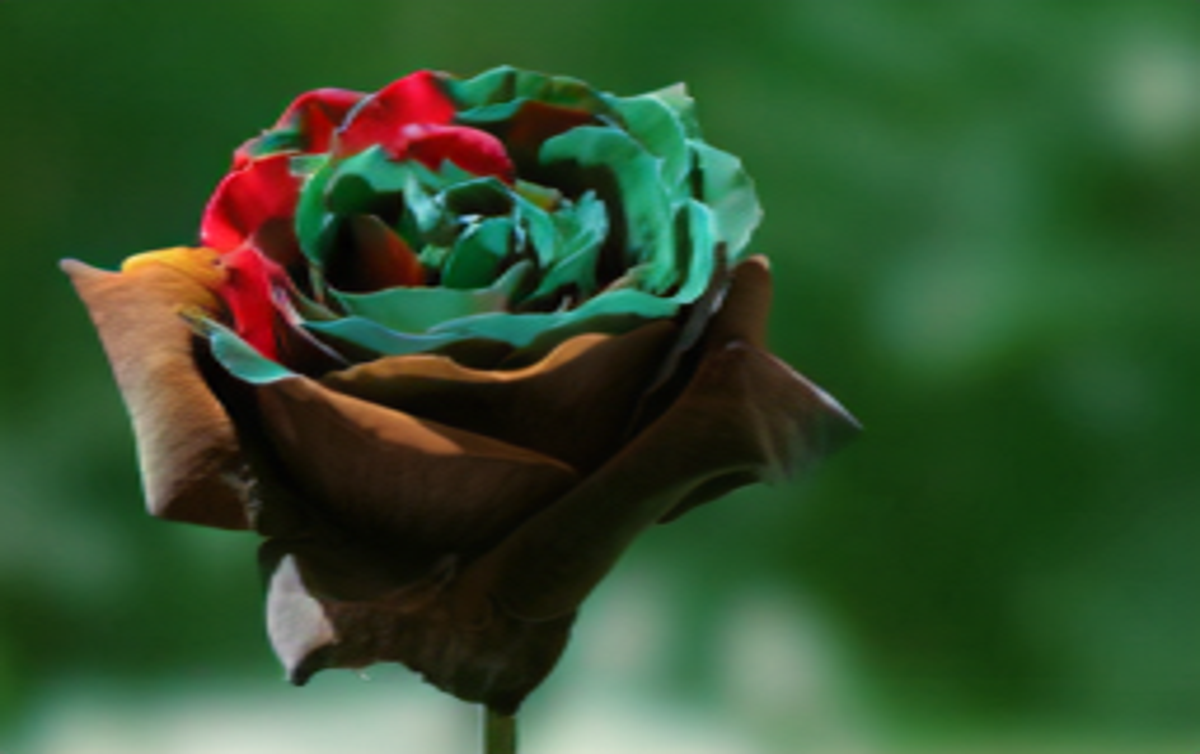} & 
            \includegraphics[width=\linewidth, height=0.81\linewidth]{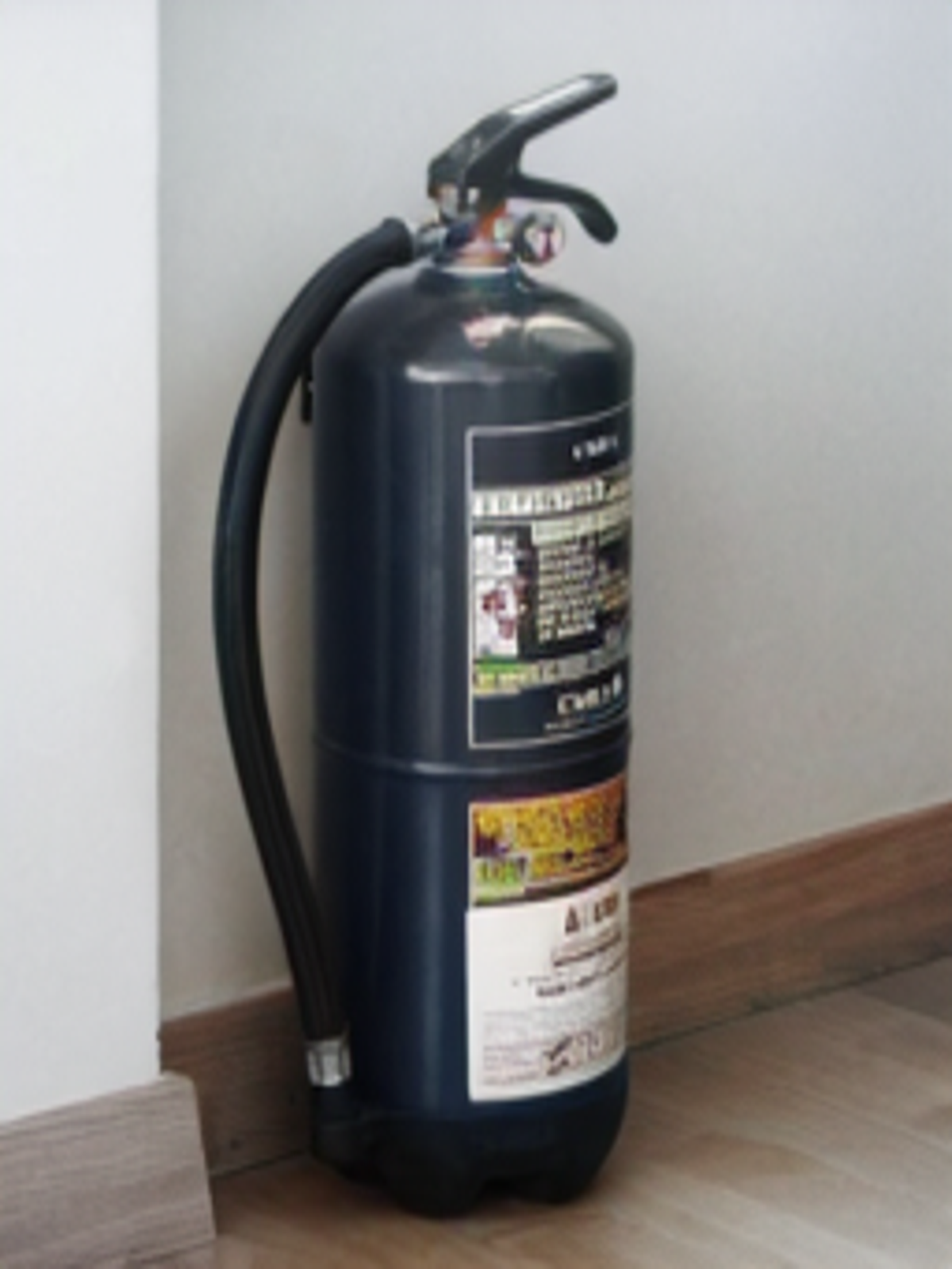} \\
    
            \scriptsize \textbf{Ours} & 
            \includegraphics[width=\linewidth, height=0.81\linewidth]{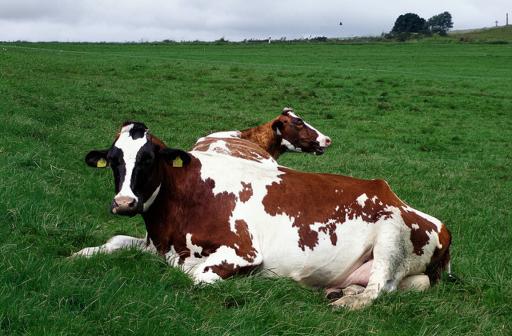} & 
            \includegraphics[width=\linewidth, height=0.81\linewidth]{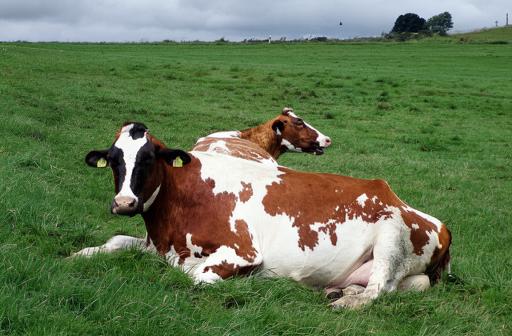} &
            \includegraphics[width=\linewidth, height=0.81\linewidth]{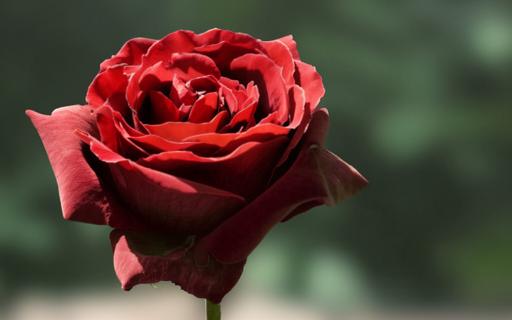} & 
            \includegraphics[width=\linewidth, height=0.81\linewidth]{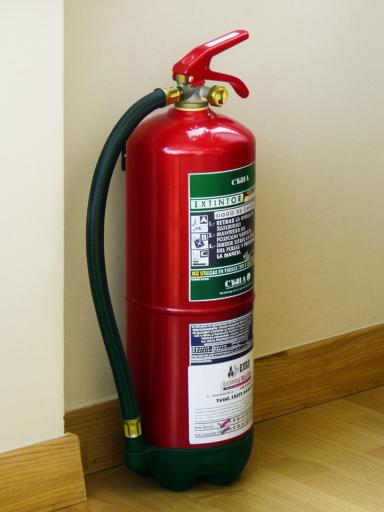} \\
        \end{tabular}
    }
    \caption{Notice that for all methods except ours, the ortho result of DDColor, and the pan result of COCO-LC, the neck of the farther-away cow is confused with the grass field. In COCO-LC and BigColor, the green color from the grass bleeds into the dark spots on the cow. Also notice that the dark colors in d. Rose and e. Fire Ext., which are supposed to be bright, are not lifted by any method other than ours. This is because the brightness of a color depends on the $L$ channel, which is fixed for the other methods. (\textbf{Zoom-in for best view})}
    \label{fig:qual2}
\end{figure*}

\begin{figure*}[htbp]
    \centering
    \setlength{\tabcolsep}{1pt}
    \renewcommand{\arraystretch}{0.5} 
    
    \begin{tabular}{c | m{0.22\linewidth} | m{0.22\linewidth} | m{0.22\linewidth} | m{0.22\linewidth} |}
        
        & \multicolumn{1}{c|}{\textbf{a. Cavalry}} & \multicolumn{1}{c|}{\textbf{b. Family}} & \multicolumn{1}{c|}{\textbf{c. Baby}} & \multicolumn{1}{c|}{\textbf{d. Group}} \\
        \cline{2-5}
        
        \scriptsize \textbf{Input} &
        \includegraphics[width=\linewidth, height=0.82\linewidth]{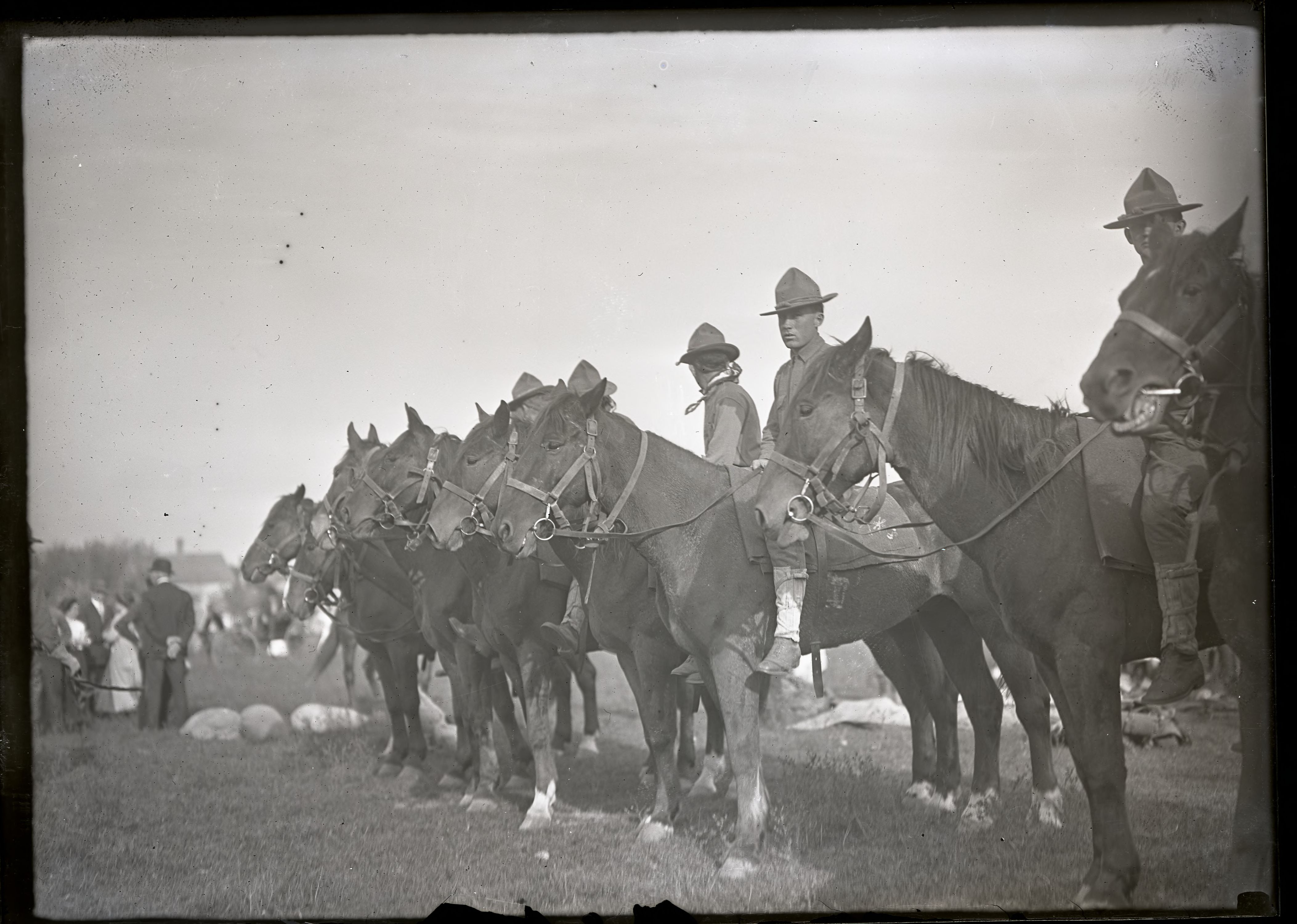} &
        \includegraphics[width=\linewidth, height=0.82\linewidth]{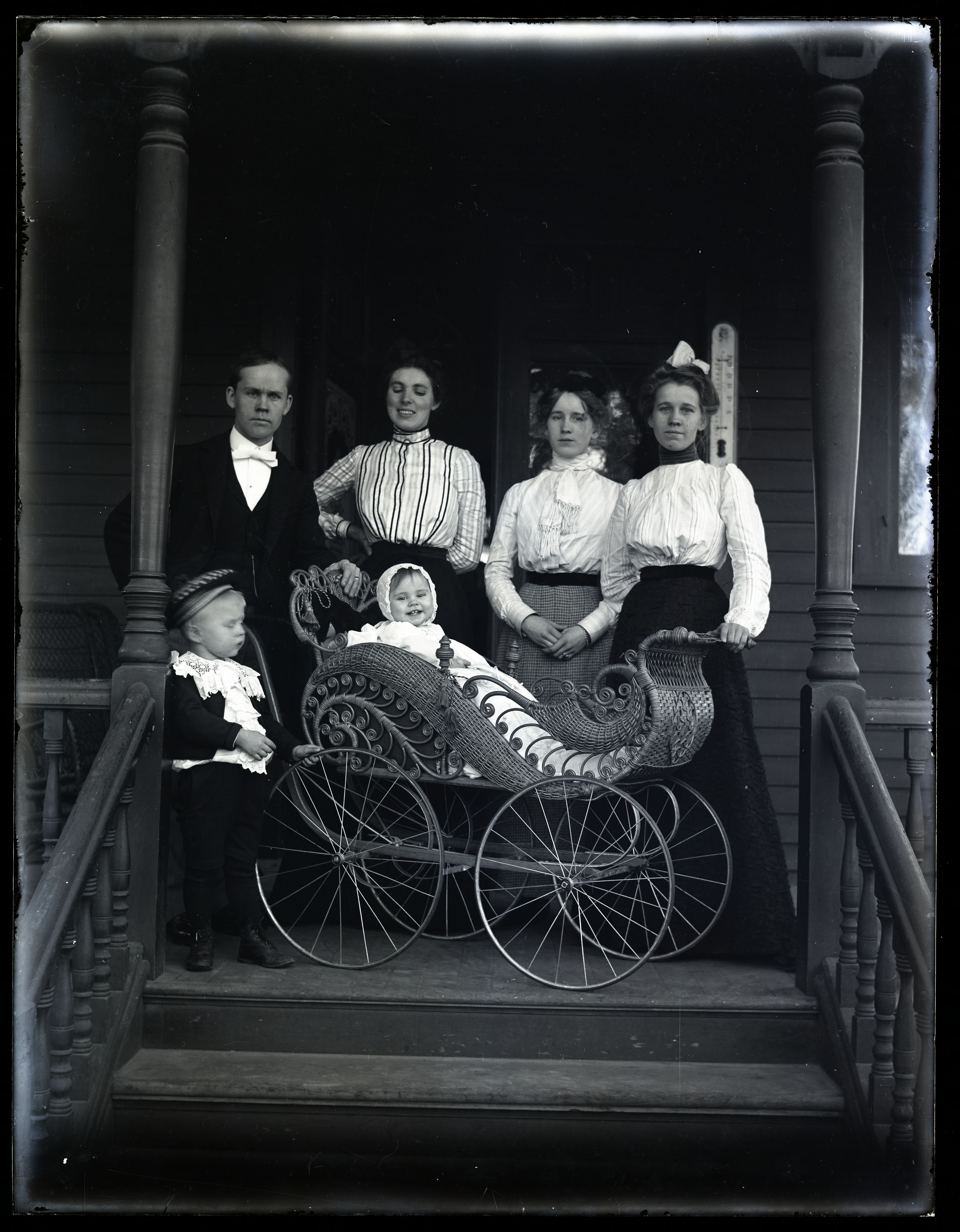} &
        \includegraphics[width=\linewidth, height=0.82\linewidth]{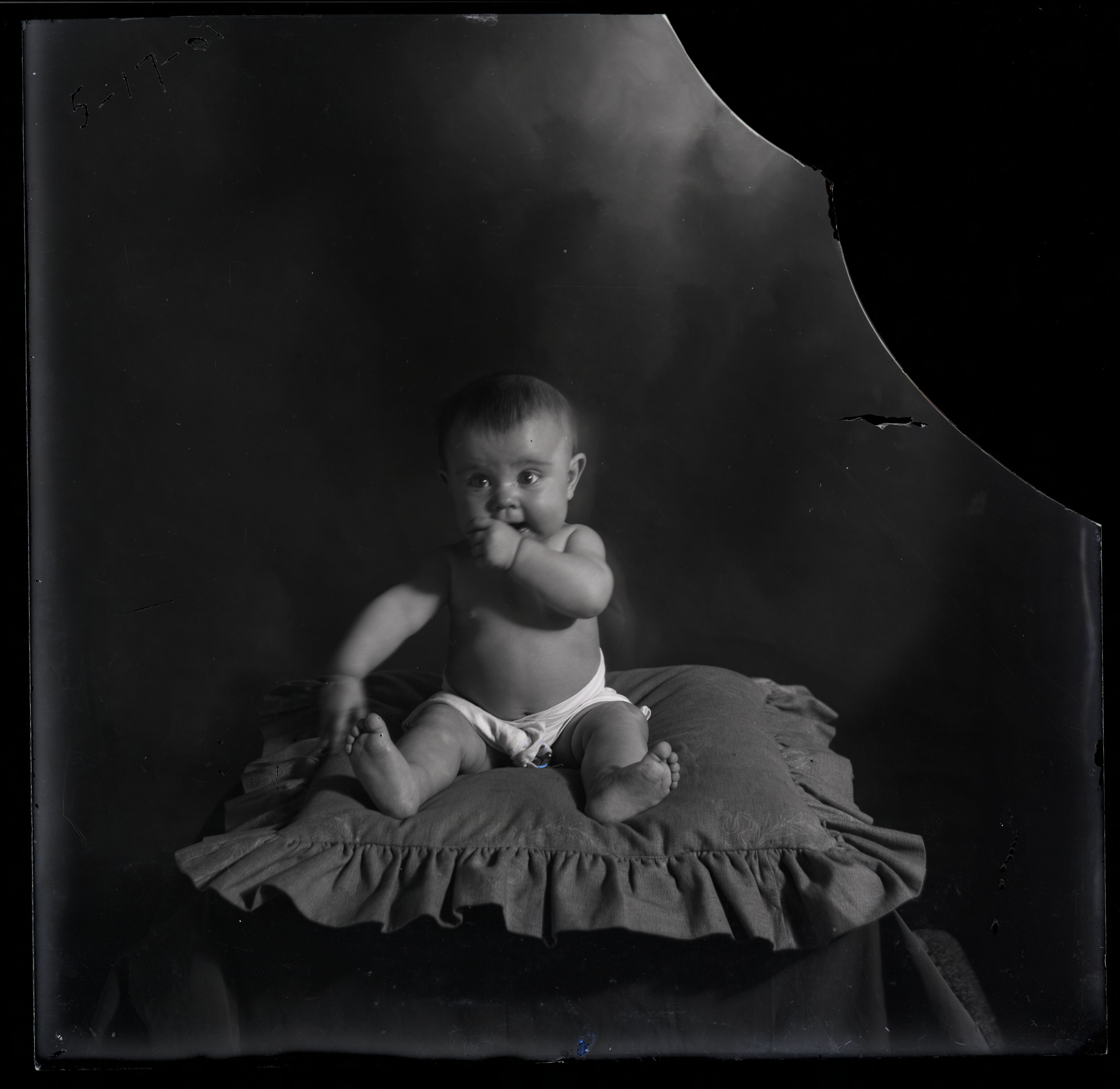} &
        \includegraphics[width=\linewidth, height=0.82\linewidth]{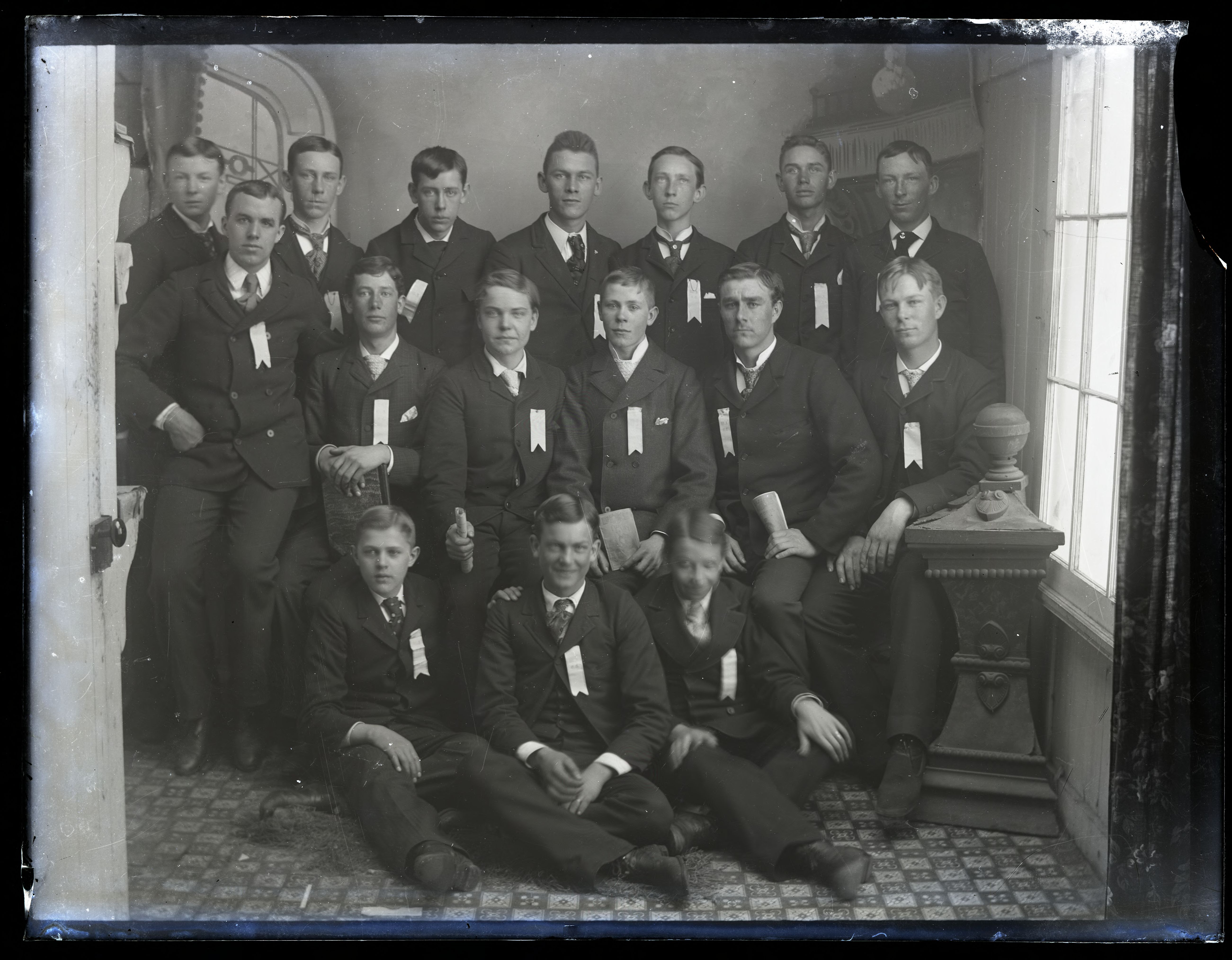} \\

        \scriptsize \textbf{DDColor} &
        \includegraphics[width=\linewidth, height=0.82\linewidth]{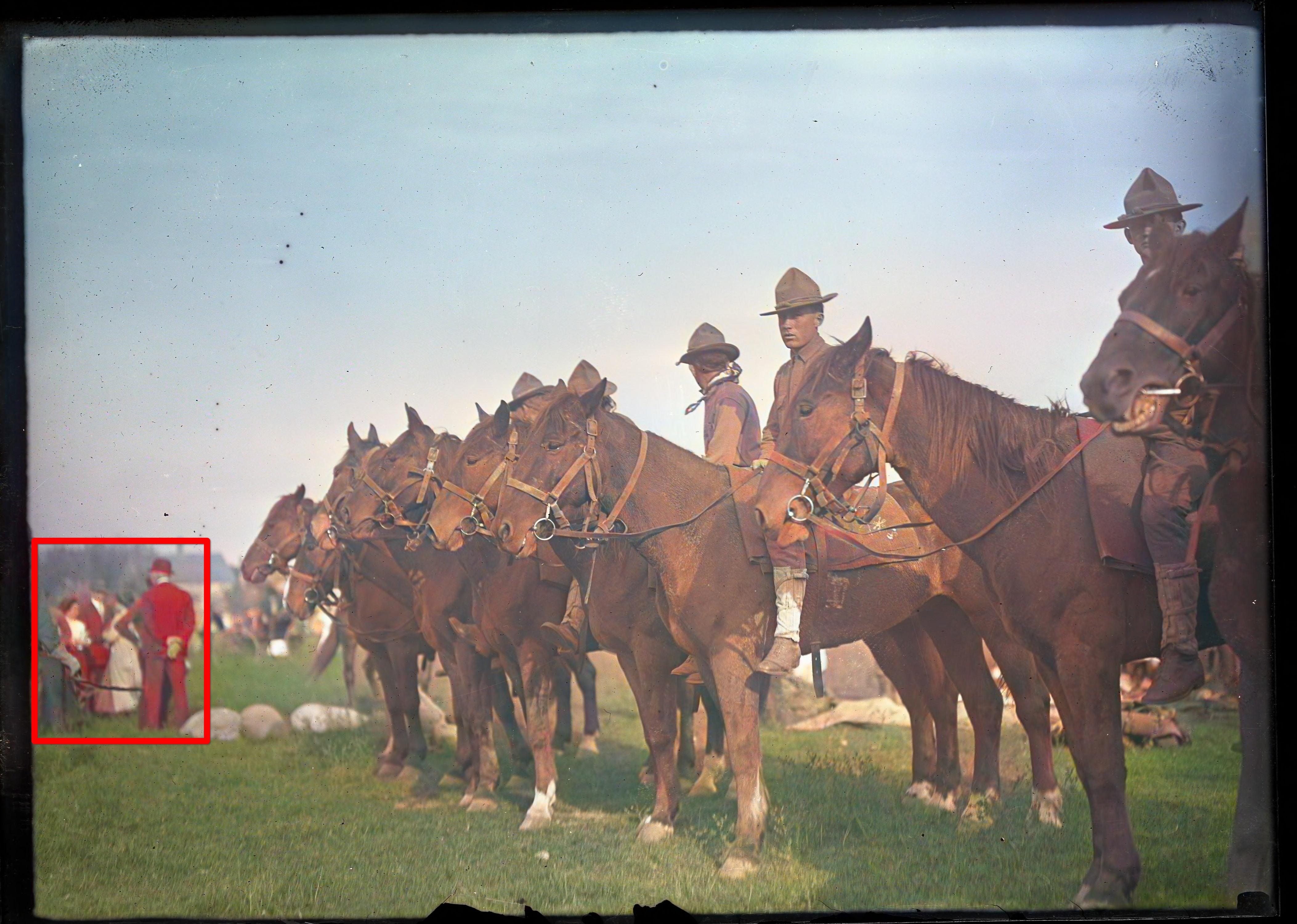} &
        \includegraphics[width=\linewidth, height=0.82\linewidth]{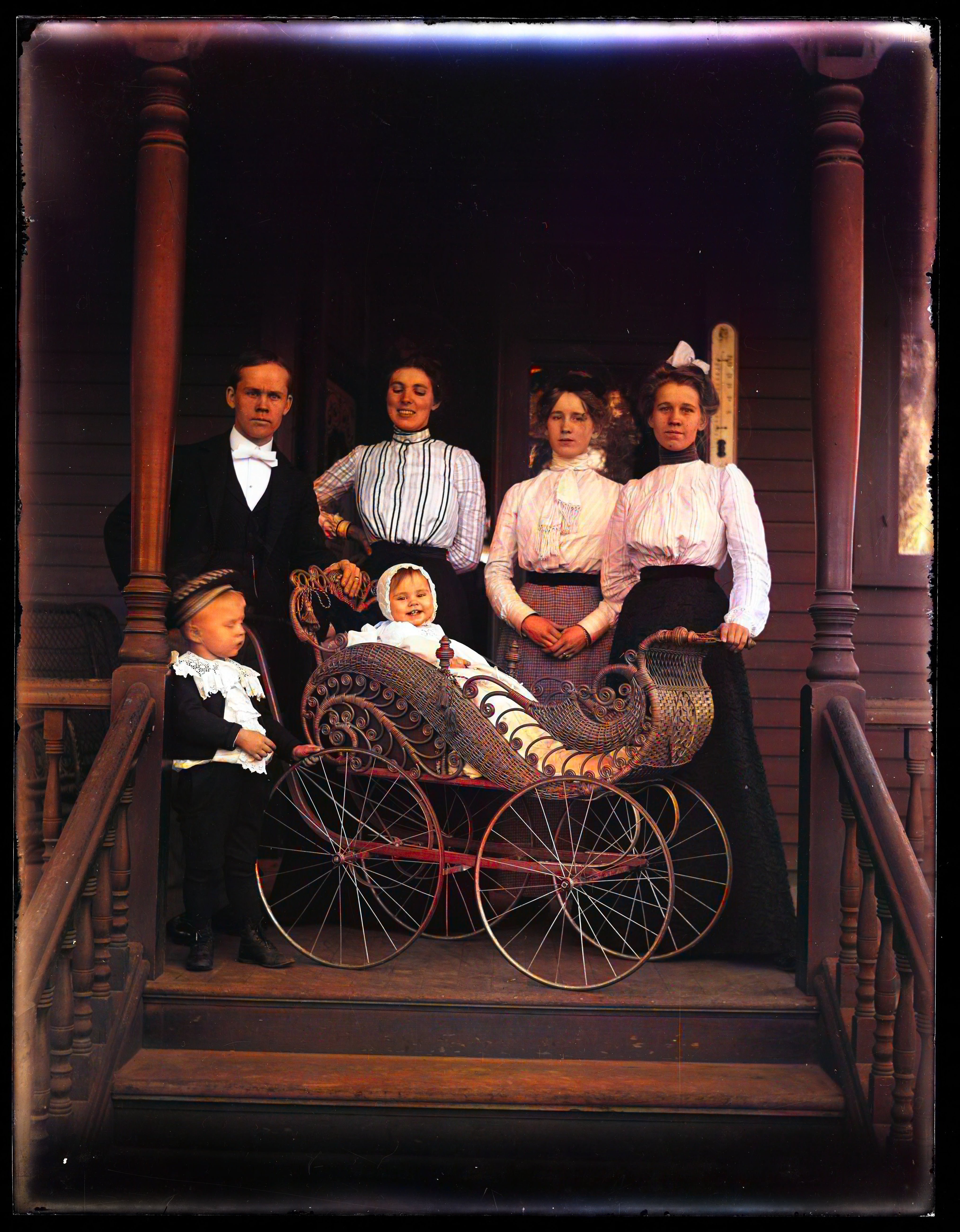} &
        \includegraphics[width=\linewidth, height=0.82\linewidth]{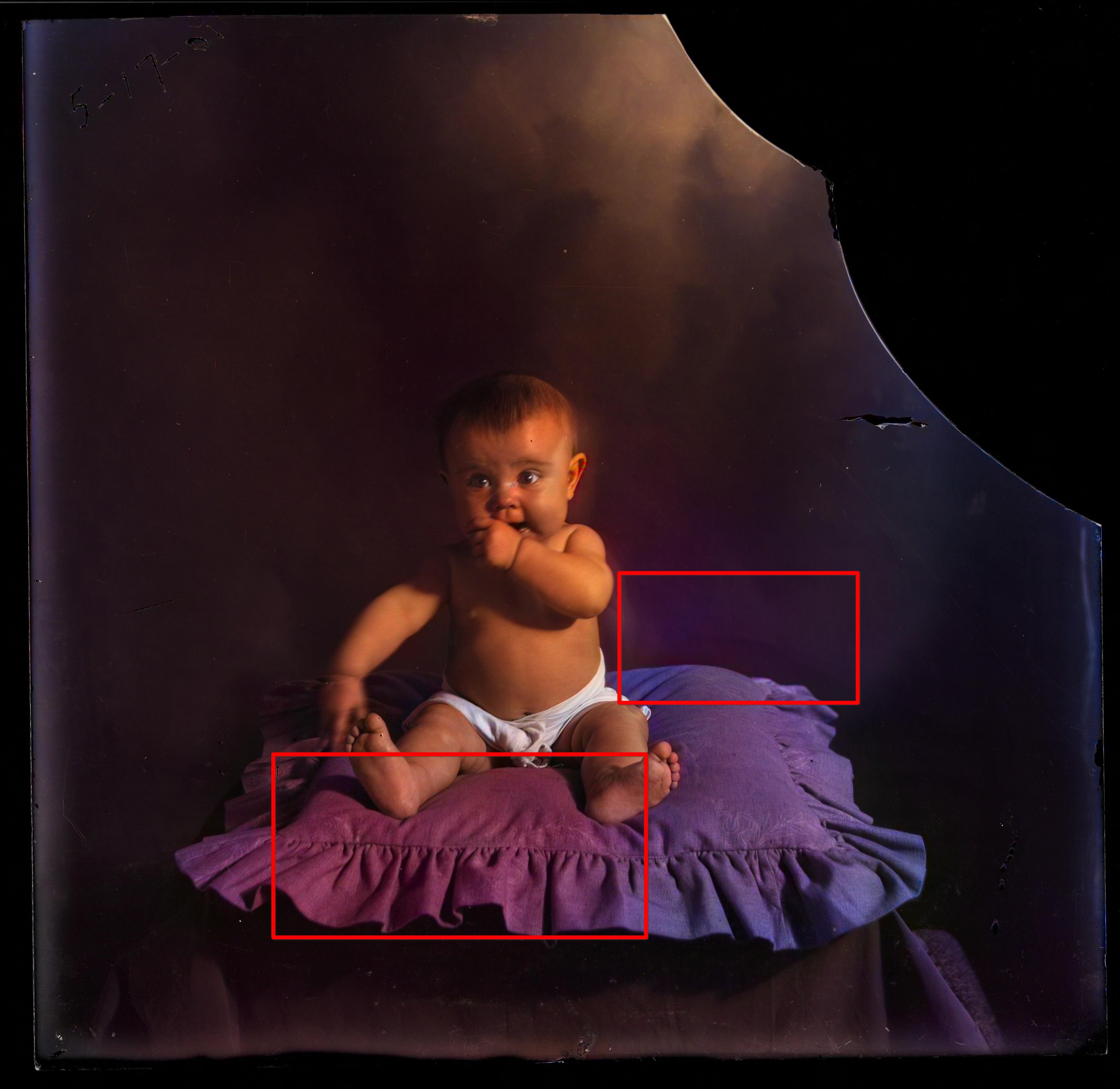} &
        \includegraphics[width=\linewidth, height=0.82\linewidth]{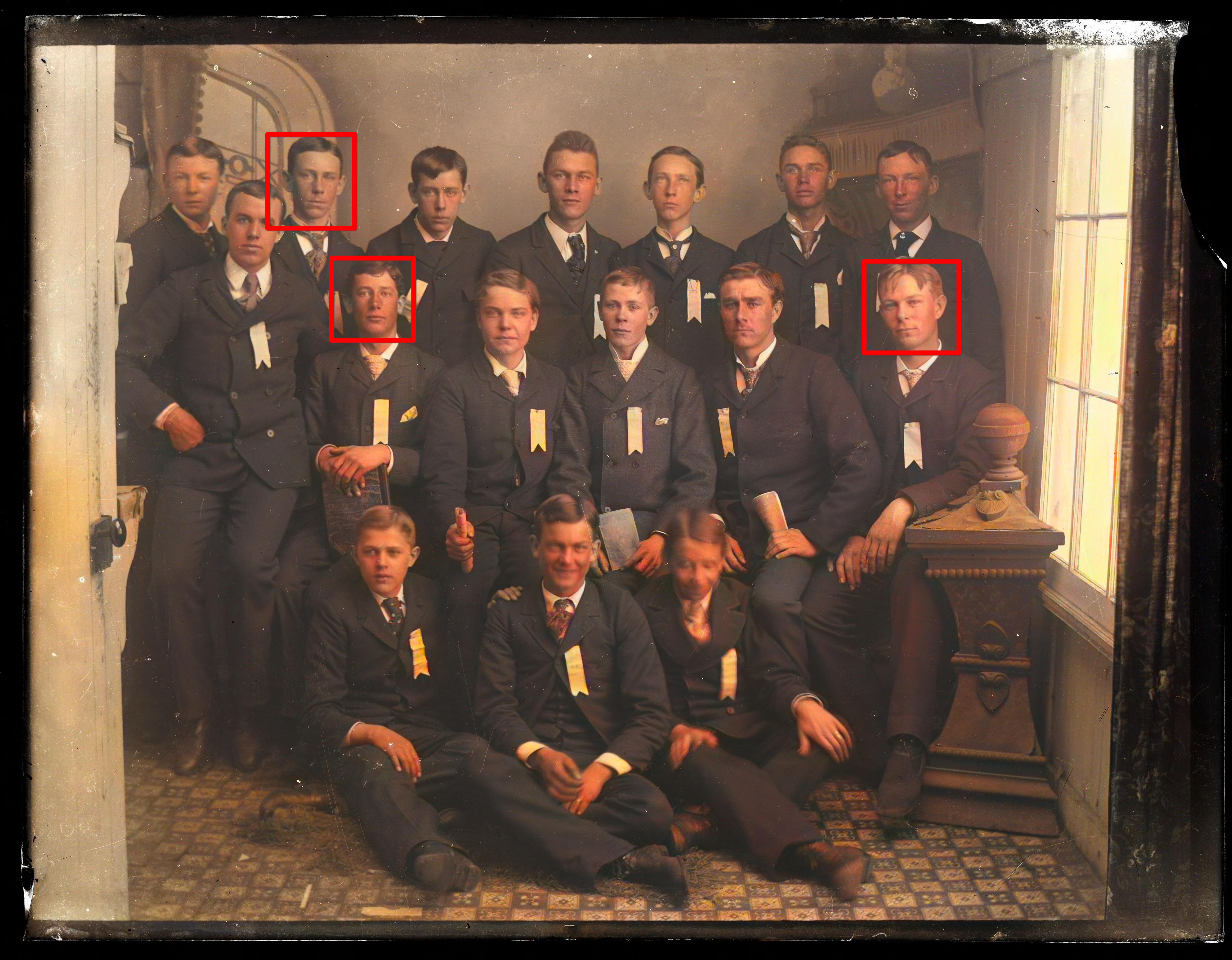} \\

        \scriptsize \textbf{DISCO} &
        \includegraphics[width=\linewidth, height=0.82\linewidth]{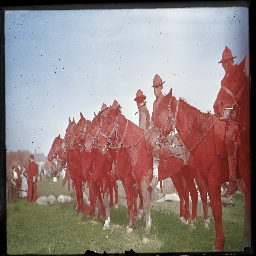} &
        \includegraphics[width=\linewidth, height=0.82\linewidth]{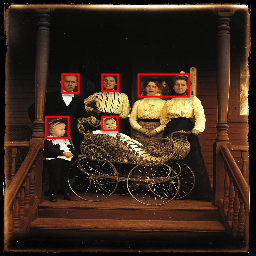} &
        \includegraphics[width=\linewidth, height=0.82\linewidth]{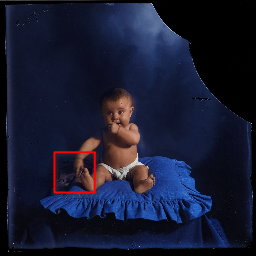} &
        \includegraphics[width=\linewidth, height=0.82\linewidth]{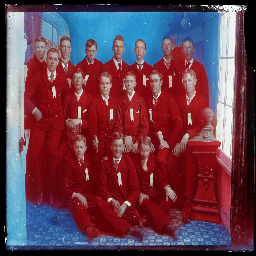} \\

        \scriptsize \textbf{COCO-LC} &
        \includegraphics[width=\linewidth, height=0.82\linewidth]{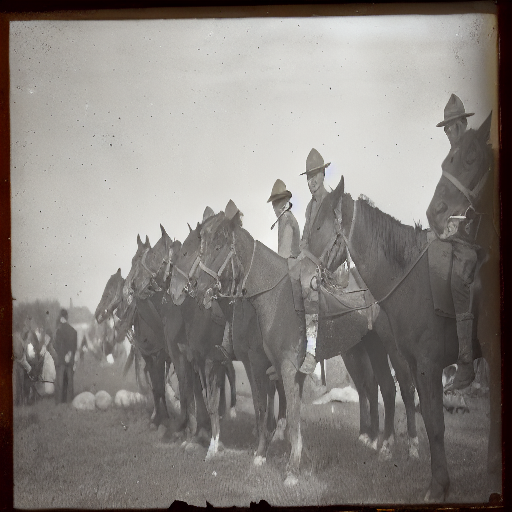} &
        \includegraphics[width=\linewidth, height=0.82\linewidth]{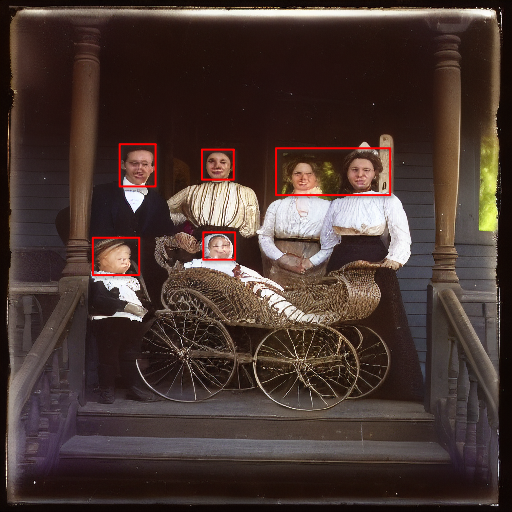} &
        \includegraphics[width=\linewidth, height=0.82\linewidth]{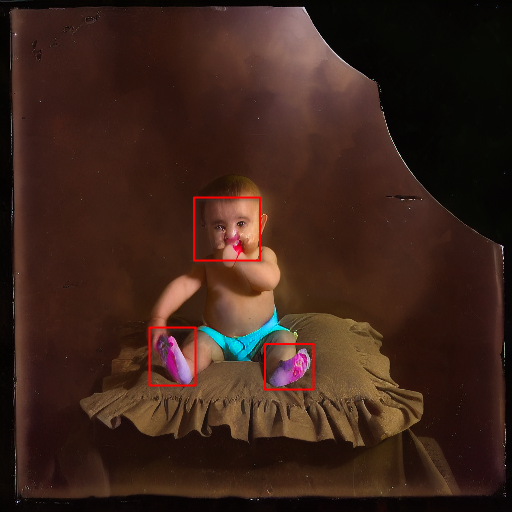} &
        \includegraphics[width=\linewidth, height=0.82\linewidth]{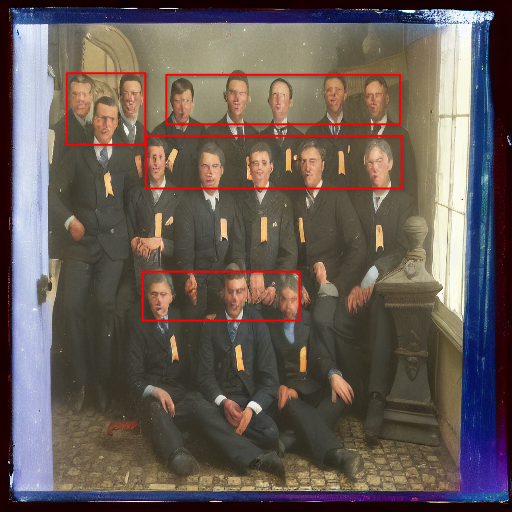} \\

        \scriptsize \textbf{BigColor} &
        \includegraphics[width=\linewidth, height=0.82\linewidth]{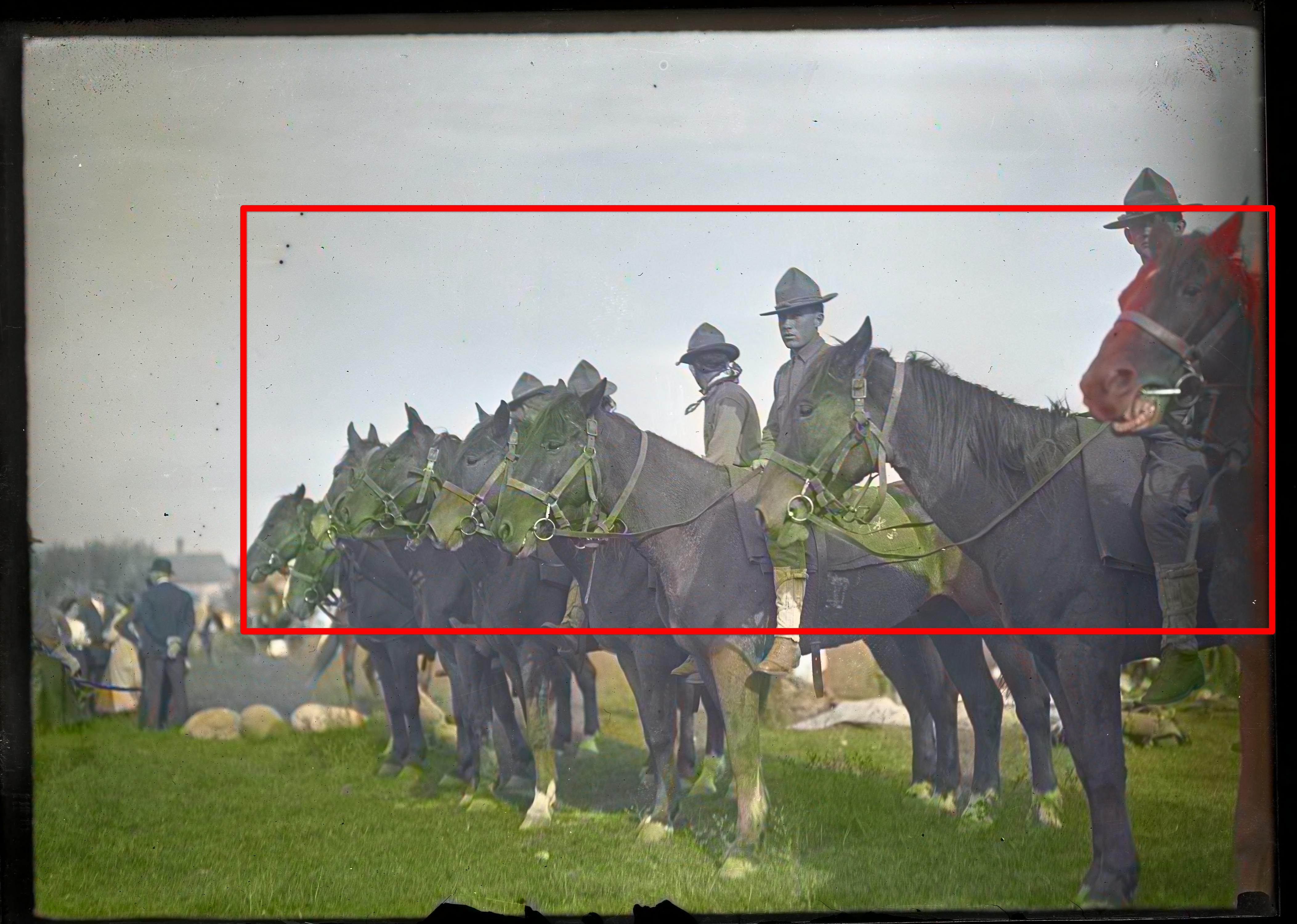} &
        \includegraphics[width=\linewidth, height=0.82\linewidth]{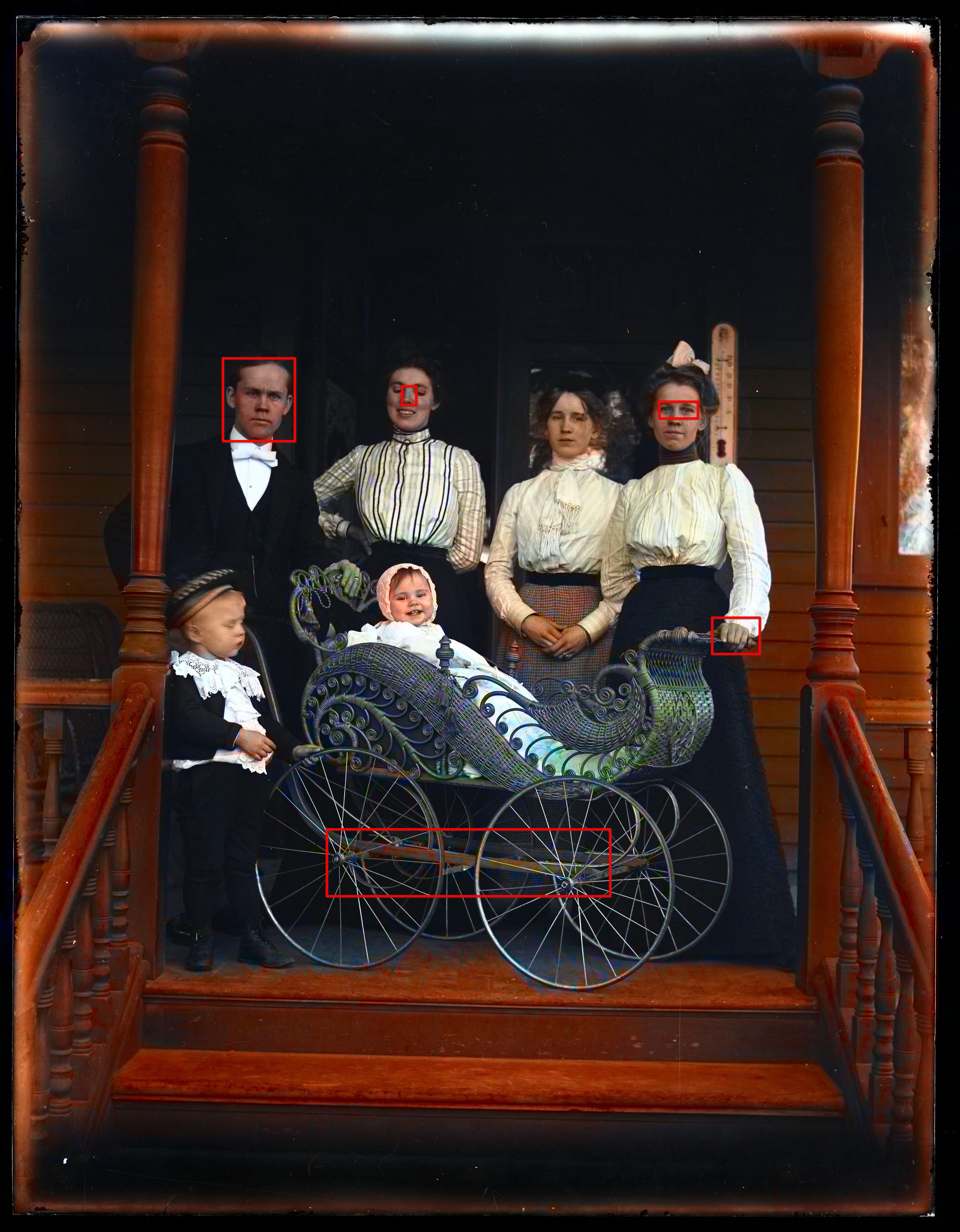} &
        \includegraphics[width=\linewidth, height=0.82\linewidth]{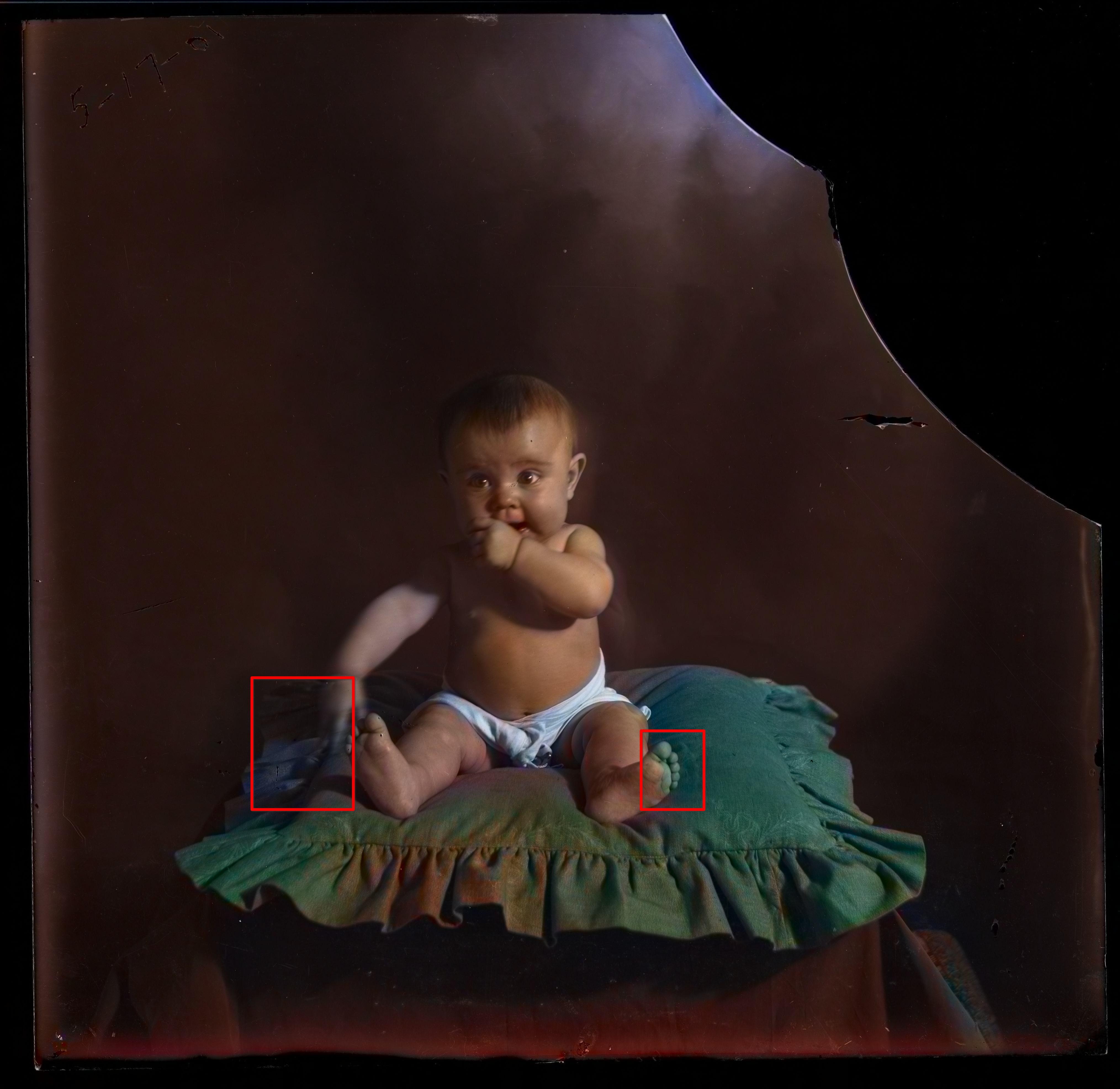} &
        \includegraphics[width=\linewidth, height=0.82\linewidth]{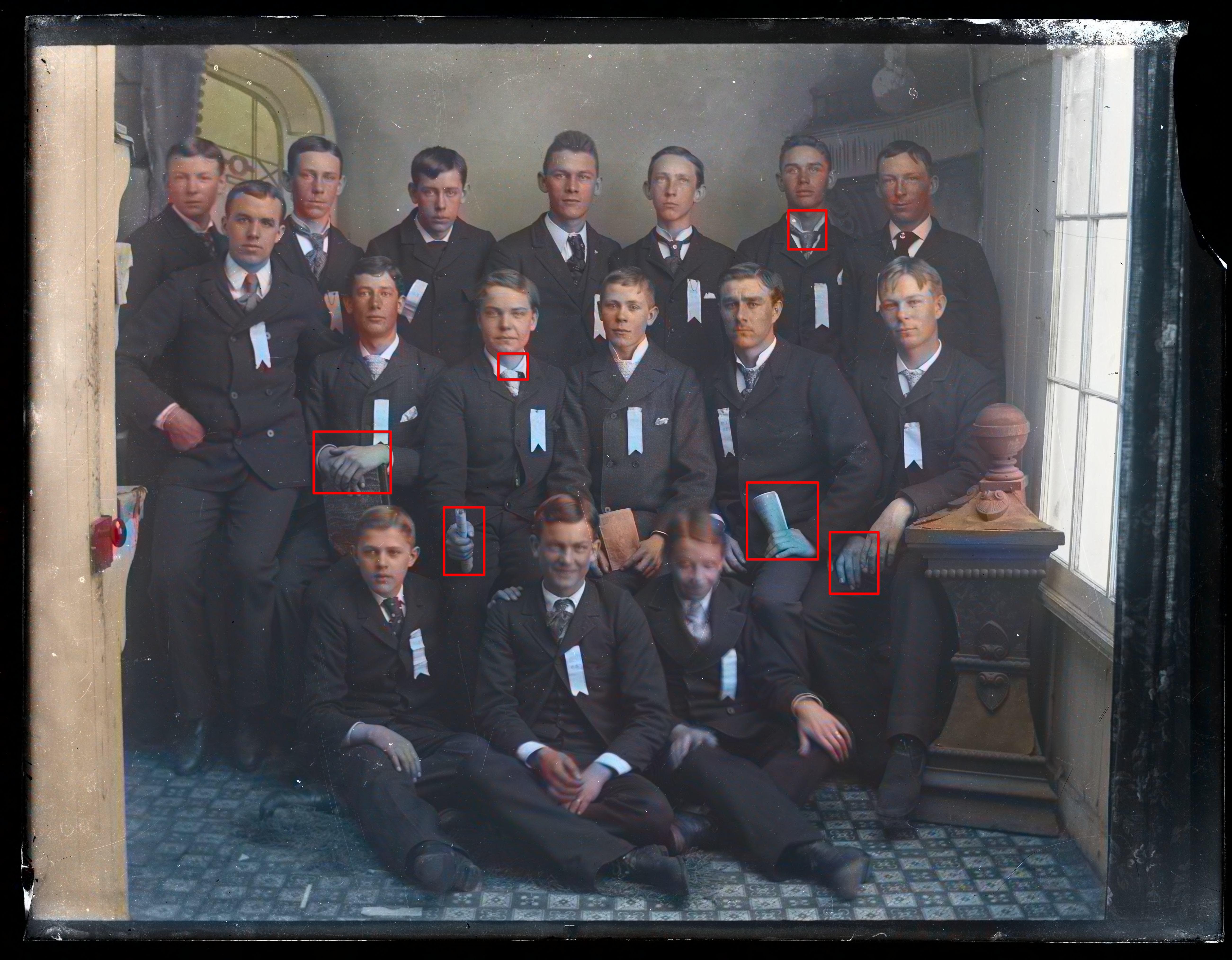} \\

        \scriptsize \textbf{UniColor} &
        \includegraphics[width=\linewidth, height=0.82\linewidth]{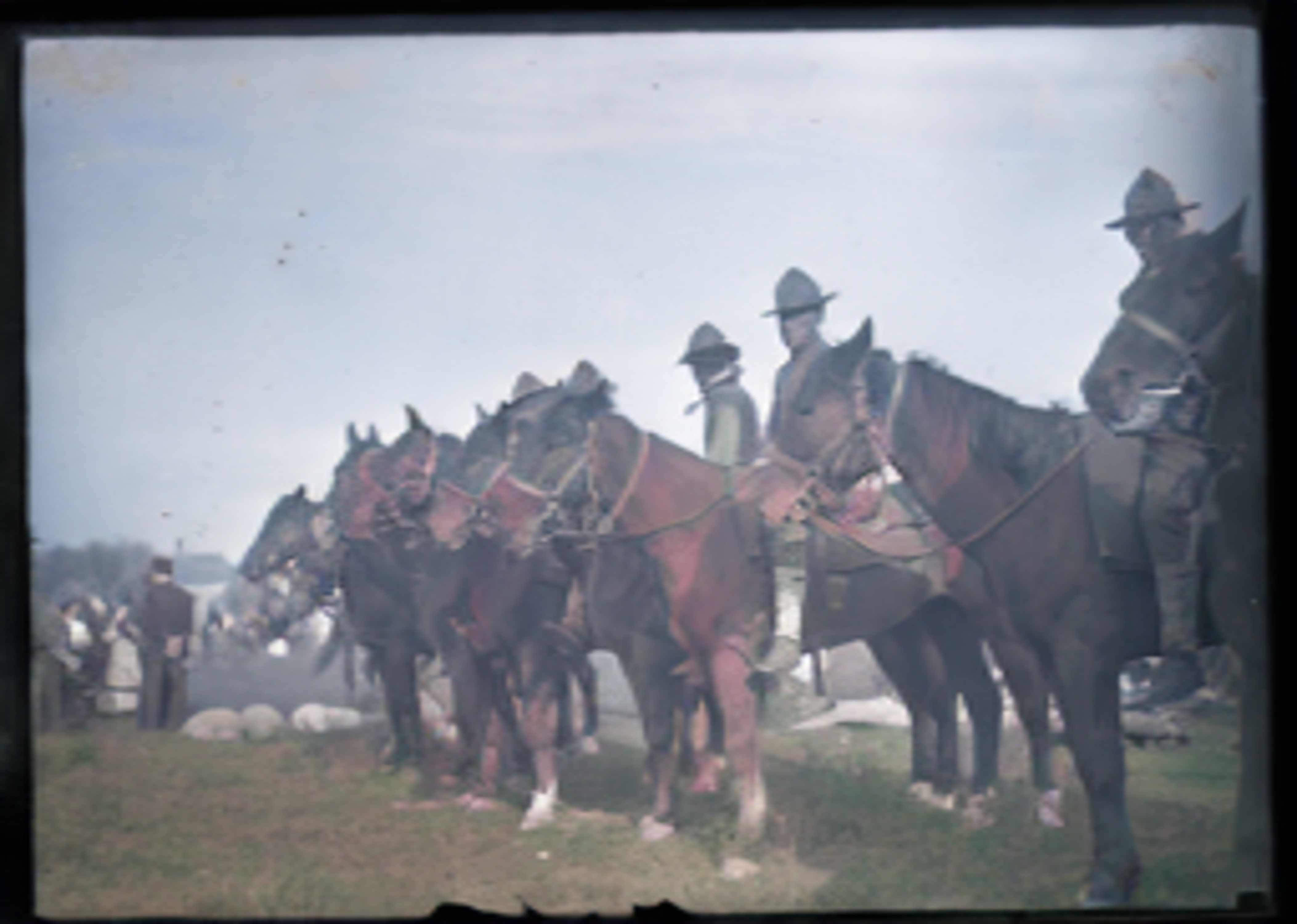} &
        \includegraphics[width=\linewidth, height=0.82\linewidth]{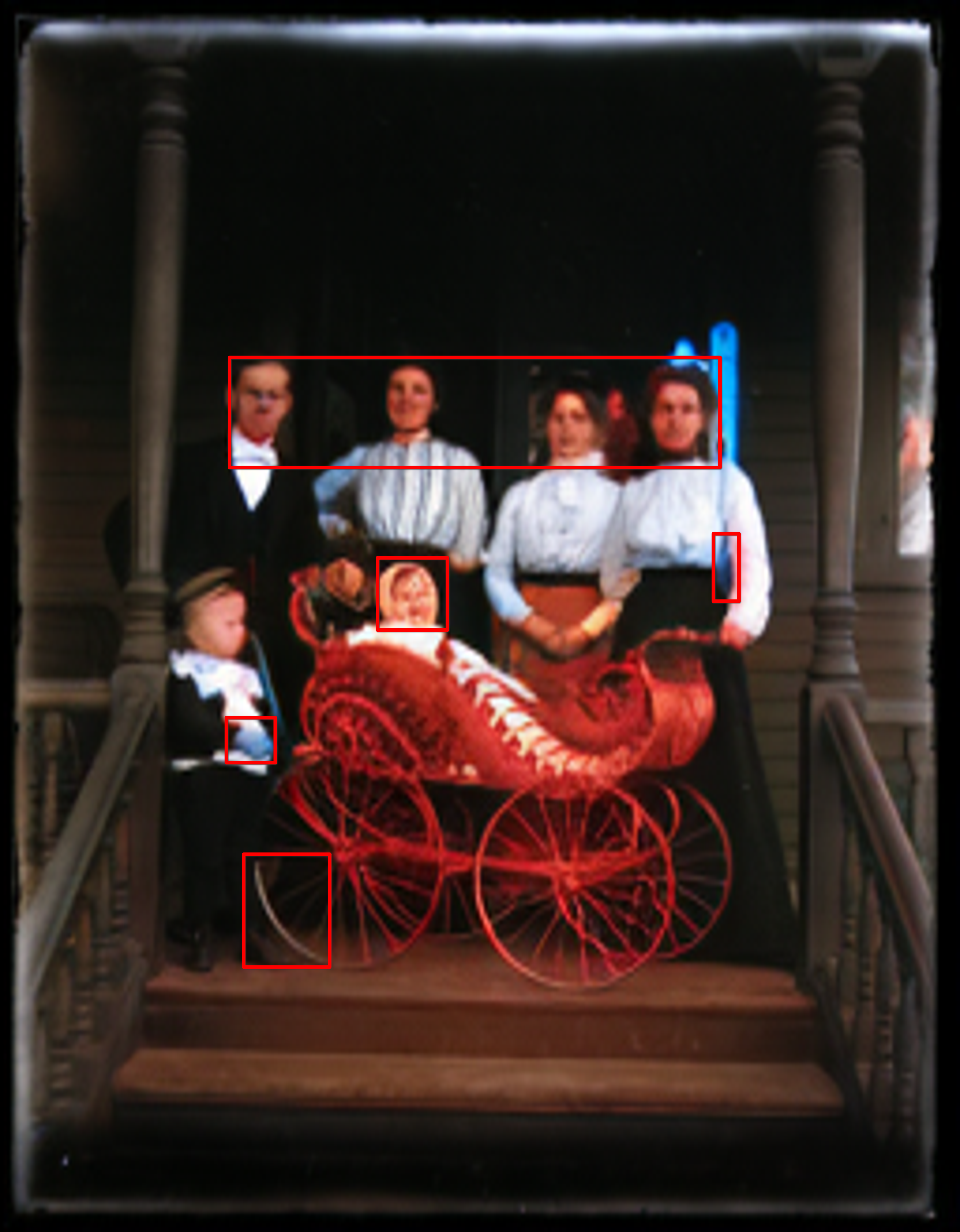} &
        \includegraphics[width=\linewidth, height=0.82\linewidth]{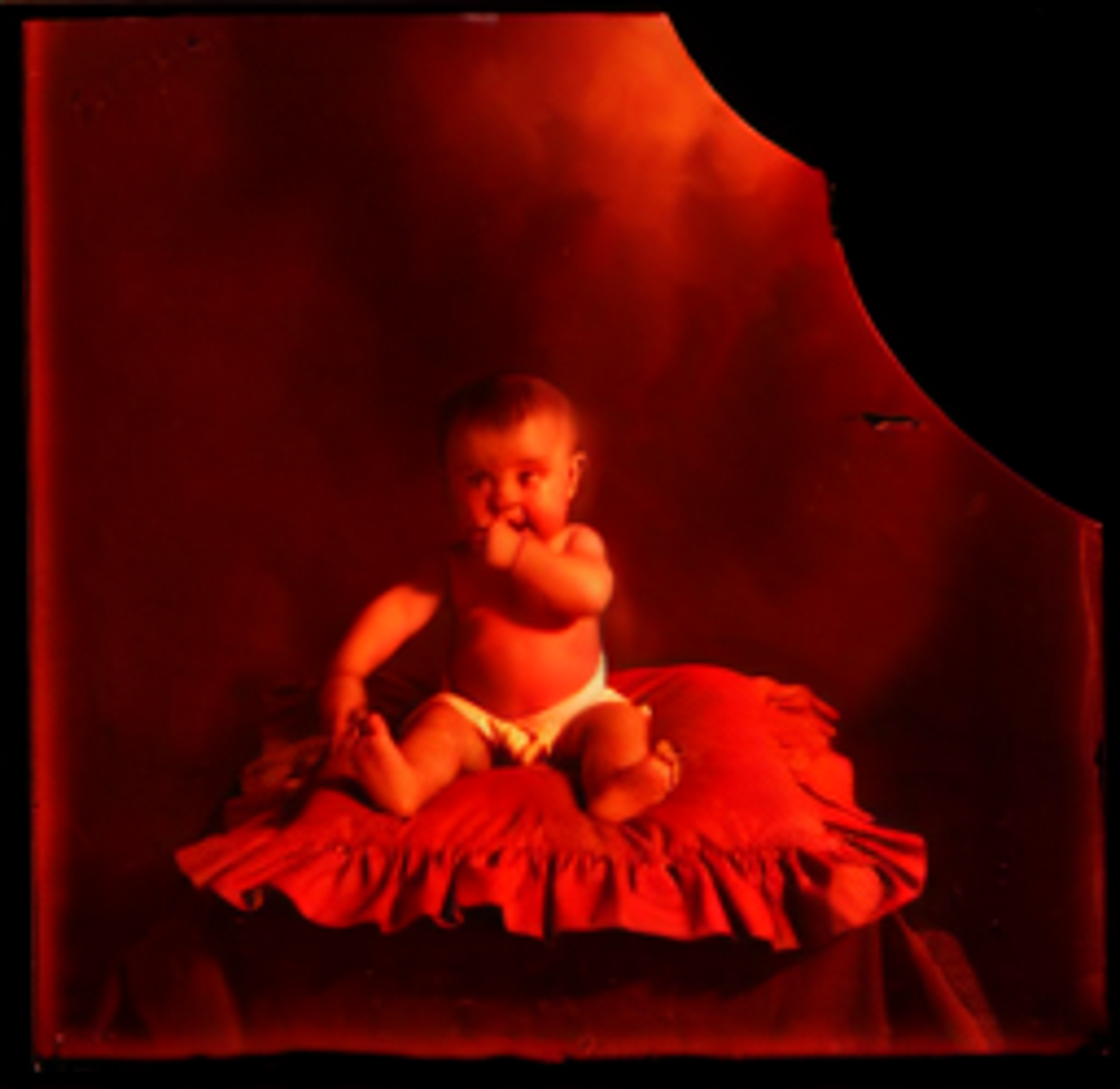} &
        \includegraphics[width=\linewidth, height=0.82\linewidth]{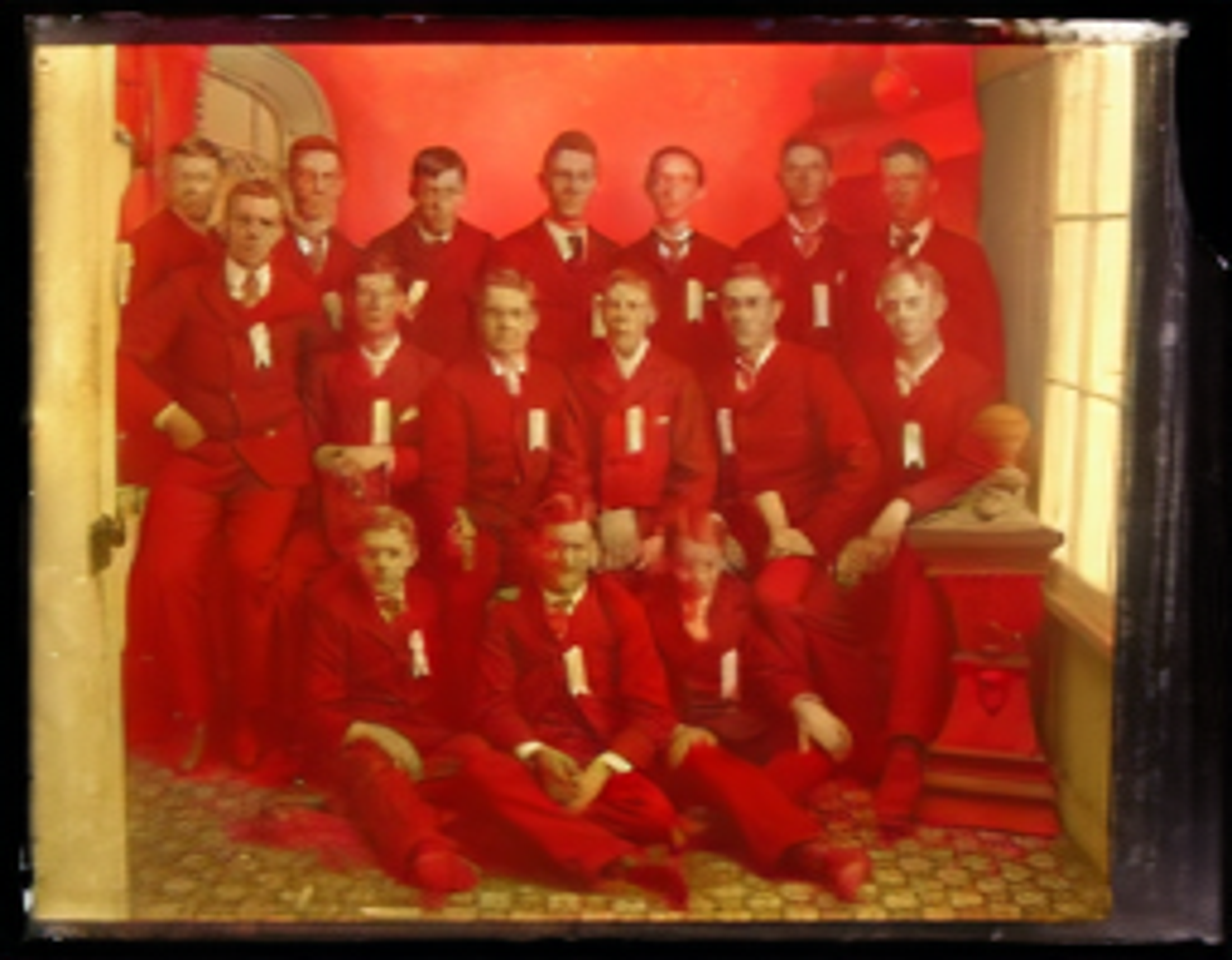} \\

        \scriptsize \textbf{Ours} &
        \includegraphics[width=\linewidth, height=0.82\linewidth]{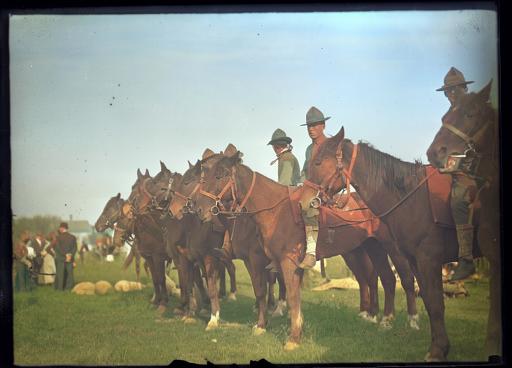} &
        \includegraphics[width=\linewidth, height=0.82\linewidth]{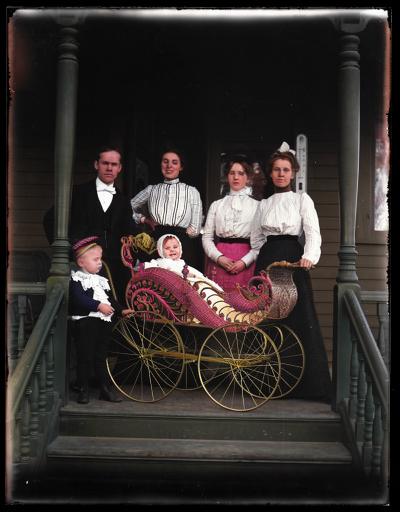} &
        \includegraphics[width=\linewidth, height=0.82\linewidth]{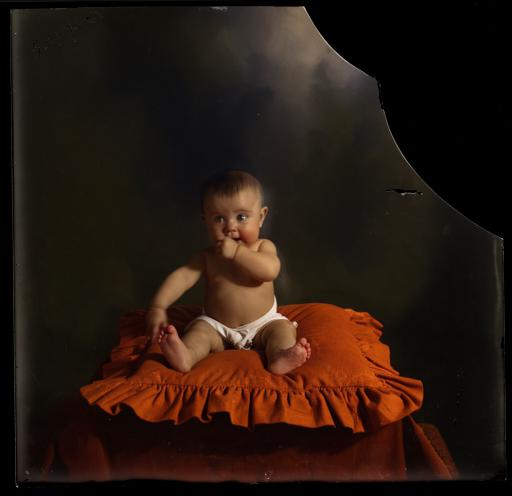} &
        \includegraphics[width=\linewidth, height=0.82\linewidth]{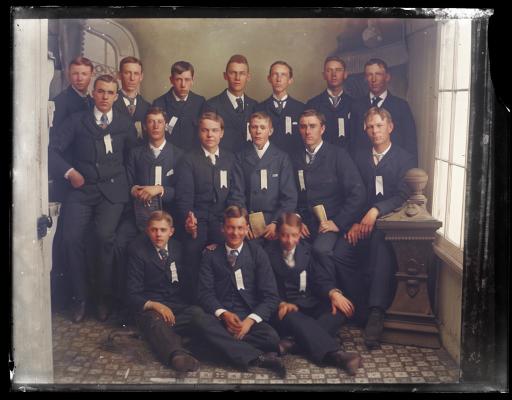} \\
        
        \cline{2-5} 
    \end{tabular}
    
    \caption{Qualitative colorization of real historical orthochromatic glass-plate photographs. Each row shows the grayscale input followed by the output of each method, and each column is a distinct archival scene: \textbf{a.} mounted cavalry, \textbf{b.} a family on a porch, \textbf{c.} a studio baby portrait, and \textbf{d.} a group of young men. Unlike our simulated-orthochromatic benchmark inputs, these are genuine red-insensitive captures, so the brightness of red regions is not preserved; our method lifts and recolors these regions more naturally than the fixed-luminance baselines. Images taken from~\cite{MDL_GlassPlates}. (\textbf{Zoom-in for best view})}
    \label{fig:qual3}
\end{figure*}

\subsection{Results}
We find that existing colorization methods degrade substantially when the input grayscale image does not preserve human-perceived luminance, as in orthochromatic imagery. In contrast, our method is substantially more robust under this shift while remaining competitive on standard grayscale inputs. This trend is reflected both quantitatively across ImageNet, COCO, and the Multi-Instance benchmark (Tables~\ref{tab:imagenet}, \ref{tab:coco}, and~\ref{tab:instance}) and qualitatively, where our outputs exhibit fewer local artifacts and more consistent color assignments (Figures~\ref{fig:qual2} and~\ref{fig:qual1}). We also evaluate on real historical orthochromatic camera images taken from~\cite{MDL_GlassPlates} (Figure~\ref{fig:qual3}). More examples are in supplementary. 

Tables~\ref{tab:coco}, \ref{tab:imagenet}, and~\ref{tab:instance} show a consistent pattern across all three benchmarks. On standard panchromatic grayscale inputs, our method remains competitive with strong recent baselines: DDColor typically achieves the best FID-family metrics, while our approach is often second-best on FID, sFID, and FID-DINO and achieves stronger perceptual color statistics such as Col-diverse and ColorNet. The advantage of our method becomes more pronounced under simulated orthochromatic inputs, where baseline performance degrades more substantially. On COCO, our method achieves the best FID-DINO and the strongest scores on all reported color metrics, while trailing DDColor only slightly on FID and sFID. On ImageNet, our method is the strongest overall in the ortho setting, achieving the best FID, sFID, FID-DINO, Colorfulness, Col-diverse, and ColorNet scores. On the Multi-Instance benchmark, our method again leads on most ortho metrics, including sFID, FID-DINO, Colorfulness, and Col-diverse, while remaining second-best on FID, Saturation, and ColorNet. Overall, these results support our central claim: the proposed luminance-agnostic formulation preserves strong performance on standard grayscale inputs and yields substantially greater robustness when grayscale formation deviates from modern panchromatic luminance.

Preserving the structure on the image during colorization: A potential concern with adapting a generative foundation model is that it may hallucinate content rather than faithfully recolor the input. To verify that our model preserves the structure of the grayscale input, we measure \emph{structural fidelity} via grayscale reprojection: we convert each colorized RGB output back to grayscale using the corresponding grayscale formation and compare it against the input grayscale image using L1, L2, and SSIM. We compare three variants of FLUX.2-klein-4B: the non-fine-tuned model (\textit{Not-FT}), a model fine-tuned only on standard panchromatic grayscale (\textit{Normal-FT}), and our full model trained with the Mixed Grayscale Objective (\textit{Mixed-FT}). Tables~\ref{tab:fidelity-coco-pixel}, \ref{tab:fidelity-imagenet-pixel}, and~\ref{tab:fidelity-multi-instance-pixel} report results on COCO, ImageNet, and Multi-Instance, respectively. The non-fine-tuned model exhibits substantially weaker structure fidelity (SSIM around $0.36$), since it tends to alter image content, whereas both fine-tuned variants preserve structure well (SSIM around $0.81$--$0.88$). This confirms that our luminance-agnostic formulation maintains the grayscale structure while still enabling full-RGB colorization.

\subsection{Human Evaluation}
To complement the automatic metrics, we conducted a human study focused on visible color artifacts. We sampled the first five images from each of the three evaluation datasets (ImageNet, COCO, and Multi-Instance) under both panchromatic and orthochromatic grayscale settings, resulting in 30 evaluation inputs per method. Each input was colorized by all compared methods and evaluated by 25 human participants.

Participants were asked a binary question: whether the displayed colorized image contained noticeable artifacts such as color bleeding, unnatural hue shifts, or spatially inconsistent colorization. We summarize the results using the \emph{Artifact-Free Rate}, defined as
\begin{equation}
\text{Artifact-Free Rate} (\%) = \frac{1}{N}\sum_{i=1}^{N} \mathbf{1}[r_i=\text{No Artifact}] \times 100,
\end{equation}
where $r_i$ denotes an individual response and $N$ is the total number of valid responses.

\begin{figure}[htbp]
     \centering
     \begin{subfigure}[b]{.45\textwidth}
         \centering
         \includegraphics[width=\linewidth]{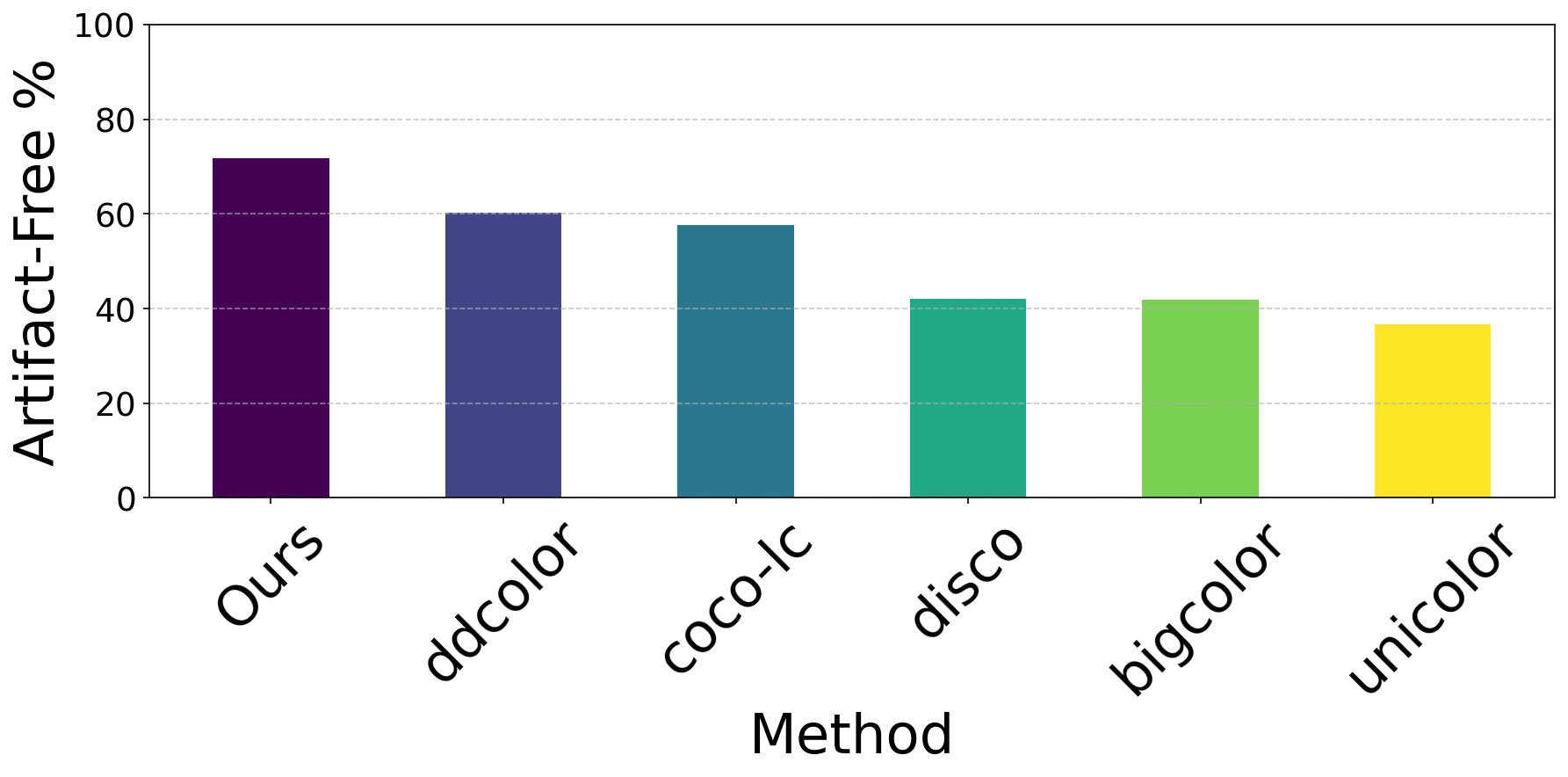}
         \caption{Artifact-Free Rate by method.}
         \label{fig:artifactreview_by_method}
     \end{subfigure}
     \hfill
     \begin{subfigure}[b]{.45\textwidth}
         \centering
         \includegraphics[width=\linewidth]{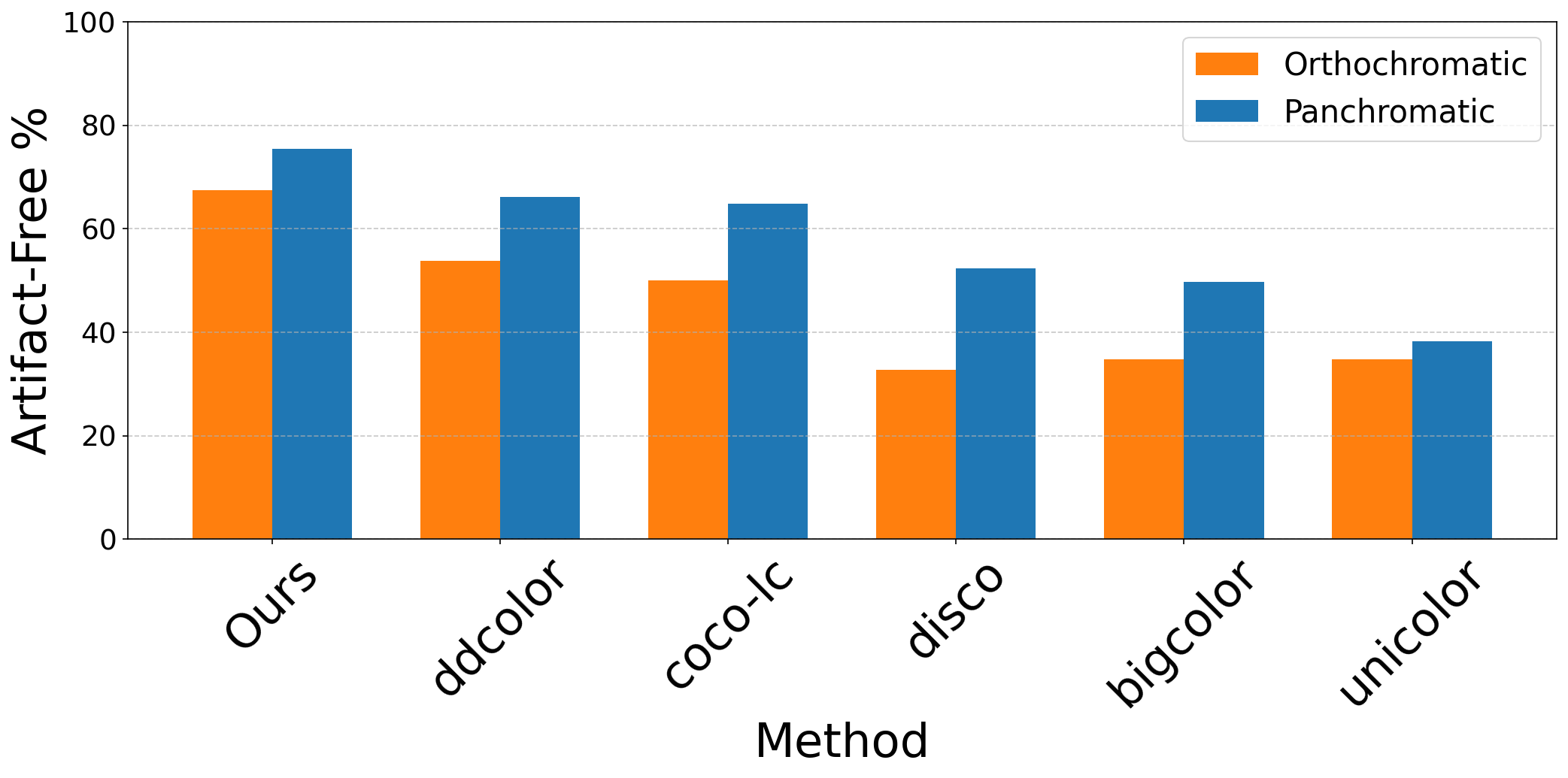} 
         \caption{Artifact-Free Rate by method and input domain.}
         \label{fig:artifactreview_by_domain}
     \end{subfigure}
        \caption{Human evaluation of visible color artifacts. We report the \emph{Artifact-Free Rate}, defined as the percentage of responses in which participants judged a colorized image to contain no noticeable artifacts. Higher values indicate that the outputs were perceived as cleaner and more visually consistent.}
\end{figure}

As shown in Figures~\ref{fig:artifactreview_by_domain} and~\ref{fig:artifactreview_by_method}, our method attains the highest Artifact-Free Rate among all compared approaches. This suggests that its outputs are perceived as more stable and less artifact-prone.

\begin{figure}[htbp]
     \centering
     \begin{subfigure}[b]{.32\textwidth}
         \centering
         \includegraphics[width=\linewidth]{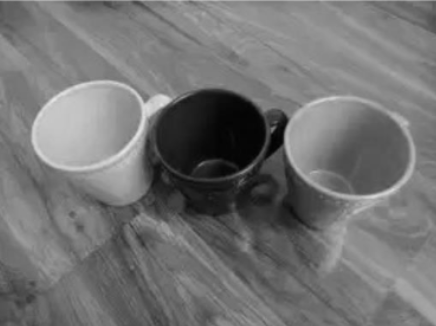}
         \caption{grayscale image}
     \end{subfigure}
     \hfill
     \begin{subfigure}[b]{.32\textwidth}
         \centering
         \includegraphics[width=\linewidth]{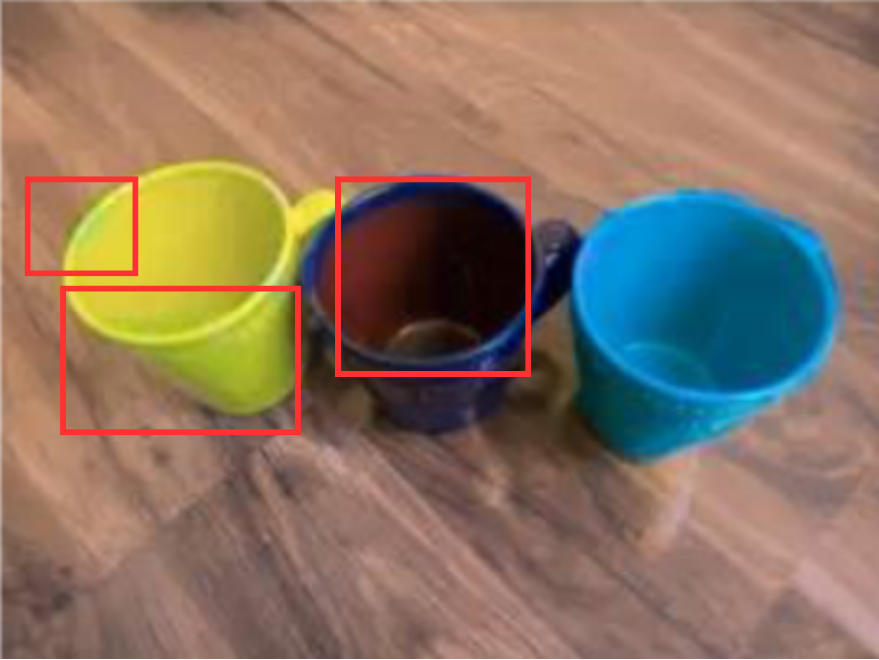} 
         \caption{COCO-LC}
     \end{subfigure}
     \hfill
        \begin{subfigure}[b]{.32\textwidth}
         \centering
         \includegraphics[width=\linewidth]{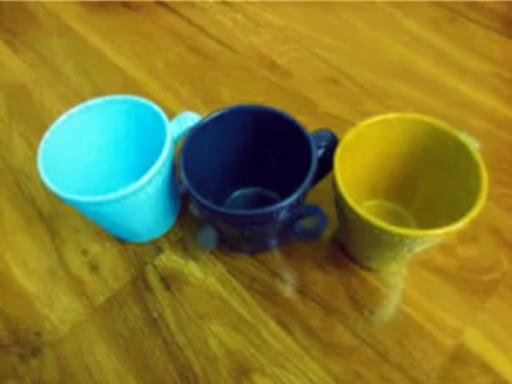} 
         \caption{Ours}
     \end{subfigure}
        \caption{a) Input grayscale image that needs to be colorized. b) The colorized output from COCO-LC. c) The colorized output from our method.
    When dealing with multiple color words in a text prompt, e.g., ``the blue cup, the yellow cup, and the cyan cup'', COCO-LC incorrectly colors the middle cup red, while our method correctly colors all three cups. Moreover, there are unnatural green spots in the yellow cup for the COCO-LC output, compared to our output, which is clean and more realistic. (\textbf{Zoom-in for best view})}
        \label{fig:cups}
\end{figure}

\textbf{Complex prompt following:}
We compare our method against COCO-LC for prompt following. We find that our method is much better at following text prompts, even very difficult ones (Figure~\ref{fig:cups}). When dealing with multiple color words in a text prompt, e.g., ``the blue cup, the yellow cup, and the cyan cup'', COCO-LC incorrectly colors the middle cup red, while our method correctly colors all three cups. More examples are given in supplementary. 

\begin{table}\centering
\scriptsize
\caption{Quantitative comparison on the COCO benchmark under standard panchromatic (\emph{pan}) and simulated orthochromatic (\emph{ortho}) grayscale inputs. The best result within each input setting (pan/ortho) is shown in bold, and the second-best is underlined.}\label{tab:coco}
\begin{tabular}{lrrrrrrrr}\toprule
& &FID $\downarrow$ &sFID $\downarrow$ &FID-DINO $\downarrow$ &Colorfulness $\uparrow$ &Col-diverse $\uparrow$ &Saturation $\uparrow$ &ColorNet $\uparrow$ \\\midrule
\multirow{2}{*}{BigColor} &pan &9.93 &10.28 &55.59 &42.53 &15.16 &17.01 &4.87 \\
&\cellcolor[HTML]{fce5cd}ortho &\cellcolor[HTML]{fce5cd}14.78 &\cellcolor[HTML]{fce5cd}15.55 &\cellcolor[HTML]{fce5cd}127.73 &\cellcolor[HTML]{fce5cd}39.4 &\cellcolor[HTML]{fce5cd}\underline{14.17} &\cellcolor[HTML]{fce5cd}15.55 &\cellcolor[HTML]{fce5cd}4.81 \\
\multirow{2}{*}{COCO-LC} &pan &8.94 &11.12 &37.83 &37.48 &14.51 &13.44 &4.87 \\
&\cellcolor[HTML]{fce5cd}ortho &\cellcolor[HTML]{fce5cd}\underline{10.45} &\cellcolor[HTML]{fce5cd}14.18 &\cellcolor[HTML]{fce5cd}53.61 &\cellcolor[HTML]{fce5cd}36.66 &\cellcolor[HTML]{fce5cd}14.16 &\cellcolor[HTML]{fce5cd}13.27 &\cellcolor[HTML]{fce5cd}\underline{4.84} \\
\multirow{2}{*}{DDColor} &pan &\textbf{7.06} &\textbf{8.26} &\textbf{32.34} &44.00 &15.50 &17.32 &\underline{5.06} \\
&\cellcolor[HTML]{fce5cd}ortho &\cellcolor[HTML]{fce5cd}\textbf{10.15} &\cellcolor[HTML]{fce5cd}\textbf{10.28} &\cellcolor[HTML]{fce5cd}\underline{47.25} &\cellcolor[HTML]{fce5cd}35.86 &\cellcolor[HTML]{fce5cd}13.02 &\cellcolor[HTML]{fce5cd}14.34 &\cellcolor[HTML]{fce5cd}4.70 \\
\multirow{2}{*}{DISCO} &pan &15.66 &11.20 &75.58 &\textbf{52.18} &\underline{16.81} &\textbf{23.77} &4.79 \\
&\cellcolor[HTML]{fce5cd}ortho &\cellcolor[HTML]{fce5cd}17.77 &\cellcolor[HTML]{fce5cd}12.31 &\cellcolor[HTML]{fce5cd}91.67 &\cellcolor[HTML]{fce5cd}\underline{41.20} &\cellcolor[HTML]{fce5cd}13.80 &\cellcolor[HTML]{fce5cd}\underline{18.11} &\cellcolor[HTML]{fce5cd}4.37 \\
\multirow{2}{*}{UniColor} &pan &11.10 &10.02 &55.70 &42.13 &14.63 &17.00 &4.78 \\
&\cellcolor[HTML]{fce5cd}ortho &\cellcolor[HTML]{fce5cd}15.22 &\cellcolor[HTML]{fce5cd}13.72 &\cellcolor[HTML]{fce5cd}91.91 &\cellcolor[HTML]{fce5cd}39.50 &\cellcolor[HTML]{fce5cd}13.81 &\cellcolor[HTML]{fce5cd}16.05 &\cellcolor[HTML]{fce5cd}4.71 \\
\multirow{2}{*}{Ours} &pan &\underline{8.73} &\underline{9.50} &\underline{37.33} &\underline{49.31} &\textbf{17.25} &\underline{19.35} &\textbf{5.14} \\
&\cellcolor[HTML]{fce5cd}ortho &\cellcolor[HTML]{fce5cd}10.67 &\cellcolor[HTML]{fce5cd}\underline{10.61} &\cellcolor[HTML]{fce5cd}\textbf{43.32} &\cellcolor[HTML]{fce5cd}\textbf{46.33} &\cellcolor[HTML]{fce5cd}\textbf{16.36} &\cellcolor[HTML]{fce5cd}\textbf{18.38} &\cellcolor[HTML]{fce5cd}\textbf{5.06} \\
\bottomrule
\end{tabular}
\end{table}

\begin{table}\centering
\scriptsize
\caption{Quantitative comparison on the ImageNet benchmark under standard panchromatic (\emph{pan}) and simulated orthochromatic (\emph{ortho}) grayscale inputs. The best result within each input setting (pan/ortho) is shown in bold, and the second-best is underlined.}\label{tab:imagenet}
\begin{tabular}{lrrrrrrrr}\toprule
& &FID $\downarrow$ &sFID $\downarrow$ &FID-DINO $\downarrow$ &Colorfulness $\uparrow$ &Col-diverse $\uparrow$ &Saturation $\uparrow$ &ColorNet $\uparrow$ \\\midrule
\multirow{2}{*}{BigColor} &pan &7.76 &8.98 &43.34 &42.45 &14.60 &18.04 &4.76 \\
&\cellcolor[HTML]{fce5cd}ortho &\cellcolor[HTML]{fce5cd}11.62 &\cellcolor[HTML]{fce5cd}13.38 &\cellcolor[HTML]{fce5cd}101.79 &\cellcolor[HTML]{fce5cd}39.15 &\cellcolor[HTML]{fce5cd}13.56 &\cellcolor[HTML]{fce5cd}16.36 &\cellcolor[HTML]{fce5cd}\underline{4.68} \\
\multirow{2}{*}{COCO-LC} &pan &9.73 &10.95 &51.22 &35.48 &13.98 &12.84 &4.72 \\
&\cellcolor[HTML]{fce5cd}ortho &\cellcolor[HTML]{fce5cd}11.88 &\cellcolor[HTML]{fce5cd}14.02 &\cellcolor[HTML]{fce5cd}79.16 &\cellcolor[HTML]{fce5cd}\underline{42.54} &\cellcolor[HTML]{fce5cd}\underline{13.91} &\cellcolor[HTML]{fce5cd}12.74 &\cellcolor[HTML]{fce5cd}4.17 \\
\multirow{2}{*}{DDColor} &pan &\textbf{5.51} &\textbf{7.44} &\textbf{27.50} &44.51 &15.31 &18.50 &\underline{4.92} \\
&\cellcolor[HTML]{fce5cd}ortho &\cellcolor[HTML]{fce5cd}\underline{8.74} &\cellcolor[HTML]{fce5cd}\underline{9.52} &\cellcolor[HTML]{fce5cd}\underline{42.36} &\cellcolor[HTML]{fce5cd}36.34 &\cellcolor[HTML]{fce5cd}12.87 &\cellcolor[HTML]{fce5cd}15.27 &\cellcolor[HTML]{fce5cd}4.56 \\
\multirow{2}{*}{DISCO} &pan &11.62 &9.52 &68.07 &\textbf{51.05} &\underline{16.17} &\textbf{24.25} &4.65 \\
&\cellcolor[HTML]{fce5cd}ortho &\cellcolor[HTML]{fce5cd}14.06 &\cellcolor[HTML]{fce5cd}10.83 &\cellcolor[HTML]{fce5cd}84.05 &\cellcolor[HTML]{fce5cd}41.07 &\cellcolor[HTML]{fce5cd}13.47 &\cellcolor[HTML]{fce5cd}\textbf{18.87} &\cellcolor[HTML]{fce5cd}4.27 \\
\multirow{2}{*}{UniColor} &pan &9.64 &8.76 &44.50 &42.18 &14.55 &17.82 &4.70 \\
&\cellcolor[HTML]{fce5cd}ortho &\cellcolor[HTML]{fce5cd}13.01 &\cellcolor[HTML]{fce5cd}12.12 &\cellcolor[HTML]{fce5cd}78.86 &\cellcolor[HTML]{fce5cd}39.83 &\cellcolor[HTML]{fce5cd}13.75 &\cellcolor[HTML]{fce5cd}17.05 &\cellcolor[HTML]{fce5cd}4.61 \\
\multirow{2}{*}{Ours} &pan &\underline{7.03} &\underline{8.48} &\underline{32.55} &\underline{48.25} &\textbf{16.88} &\underline{19.54} &\textbf{5.03} \\
&\cellcolor[HTML]{fce5cd}ortho &\cellcolor[HTML]{fce5cd}\textbf{7.91} &\cellcolor[HTML]{fce5cd}\textbf{9.35} &\cellcolor[HTML]{fce5cd}\textbf{38.48} &\cellcolor[HTML]{fce5cd}\textbf{45.59} &\cellcolor[HTML]{fce5cd}\textbf{16.06} &\cellcolor[HTML]{fce5cd}\underline{18.74} &\cellcolor[HTML]{fce5cd}\textbf{4.98} \\
\bottomrule
\end{tabular}
\end{table}

\begin{table}\centering
\scriptsize
\caption{Quantitative comparison on the Multi-Instance benchmark under standard grayscale (\emph{pan}) and simulated orthochromatic (\emph{ortho}) inputs. The best result within each input setting (pan/ortho) is shown in bold, and the second-best is underlined.}\label{tab:instance}
\begin{tabular}{lrrrrrrrr}\toprule
& &FID $\downarrow$ &sFID $\downarrow$ &FID-DINO $\downarrow$ &Colorfulness $\uparrow$ &Col-diverse $\uparrow$ &Saturation $\uparrow$ &ColorNet $\uparrow$ \\\midrule
\multirow{2}{*}{BigColor} &pan &8.37 &7.97 &51.48 &44.17 &16.13 &16.88 &5.05 \\
&\cellcolor[HTML]{fce5cd}ortho &\cellcolor[HTML]{fce5cd}13.11 &\cellcolor[HTML]{fce5cd}12.36 &\cellcolor[HTML]{fce5cd}118.92 &\cellcolor[HTML]{fce5cd}40.08 &\cellcolor[HTML]{fce5cd}14.71 &\cellcolor[HTML]{fce5cd}15.32 &\cellcolor[HTML]{fce5cd}4.91 \\
\multirow{2}{*}{COCO-LC} &pan &\underline{6.69} &10.12 &47.86 &39.46 &15.37 &13.52 &5.07 \\
&\cellcolor[HTML]{fce5cd}ortho &\cellcolor[HTML]{fce5cd}\textbf{7.79} &\cellcolor[HTML]{fce5cd}11.15 &\cellcolor[HTML]{fce5cd}45.38 &\cellcolor[HTML]{fce5cd}40.60 &\cellcolor[HTML]{fce5cd}\underline{15.99} &\cellcolor[HTML]{fce5cd}13.72 &\cellcolor[HTML]{fce5cd}\textbf{5.21} \\
\multirow{2}{*}{DDColor} &pan &\textbf{5.09} &\textbf{6.22} &\textbf{26.76} &45.46 &16.31 &17.15 &\textbf{5.26} \\
&\cellcolor[HTML]{fce5cd}ortho &\cellcolor[HTML]{fce5cd}8.33 &\cellcolor[HTML]{fce5cd}\underline{7.93} &\cellcolor[HTML]{fce5cd}\underline{40.46} &\cellcolor[HTML]{fce5cd}36.36 &\cellcolor[HTML]{fce5cd}13.48 &\cellcolor[HTML]{fce5cd}14.00 &\cellcolor[HTML]{fce5cd}4.84 \\
\multirow{2}{*}{DISCO} &pan &13.15 &8.56 &63.81 &\textbf{52.94} &\textbf{17.55} &\textbf{23.01} &4.87 \\
&\cellcolor[HTML]{fce5cd}ortho &\cellcolor[HTML]{fce5cd}15.07 &\cellcolor[HTML]{fce5cd}9.49 &\cellcolor[HTML]{fce5cd}77.98 &\cellcolor[HTML]{fce5cd}\underline{41.39} &\cellcolor[HTML]{fce5cd}14.28 &\cellcolor[HTML]{fce5cd}\textbf{17.42} &\cellcolor[HTML]{fce5cd}4.42 \\
\multirow{2}{*}{UniColor} &pan &8.86 &7.55 &47.17 &43.57 &15.33 &16.93 &4.92 \\
&\cellcolor[HTML]{fce5cd}ortho &\cellcolor[HTML]{fce5cd}12.46 &\cellcolor[HTML]{fce5cd}10.72 &\cellcolor[HTML]{fce5cd}77.46 &\cellcolor[HTML]{fce5cd}40.49 &\cellcolor[HTML]{fce5cd}14.43 &\cellcolor[HTML]{fce5cd}15.87 &\cellcolor[HTML]{fce5cd}4.81 \\
\multirow{2}{*}{Ours} &pan &7.14 &\underline{7.02} &\underline{30.94} &\underline{48.64} &\underline{17.45} &\underline{18.26} &\underline{5.21} \\
&\cellcolor[HTML]{fce5cd}ortho &\cellcolor[HTML]{fce5cd}\underline{8.02} &\cellcolor[HTML]{fce5cd}\textbf{7.88} &\cellcolor[HTML]{fce5cd}\textbf{35.71} &\cellcolor[HTML]{fce5cd}\textbf{45.22} &\cellcolor[HTML]{fce5cd}\textbf{16.37} &\cellcolor[HTML]{fce5cd}\underline{17.16} &\cellcolor[HTML]{fce5cd}\underline{5.11} \\
\bottomrule
\end{tabular}
\end{table}

\begin{table}[ht]
\centering
\scriptsize
\caption{Structural fidelity on \textbf{COCO} via grayscale reprojection: each colorized output is converted back to grayscale and compared to the input grayscale image. L1 and L2: lower is better; SSIM: higher is better. Best in bold.}
\label{tab:fidelity-coco-pixel}
\setlength{\tabcolsep}{5pt}
\begin{tabular}{l ccc ccc}
\toprule
\multirow{2}{*}{Method} & \multicolumn{3}{c}{\textbf{Ortho}} & \multicolumn{3}{c}{\textbf{Pan}} \\
\cmidrule(lr){2-4} \cmidrule(lr){5-7}
& L1 $\downarrow$ & L2 $\downarrow$ & SSIM $\uparrow$ & L1 $\downarrow$ & L2 $\downarrow$ & SSIM $\uparrow$ \\
\midrule
Not-FT          & 0.136 & 0.301 & 0.360 & 0.139 & 0.308 & 0.354 \\
Normal-FT       & \textbf{0.050} & \textbf{0.164} & \textbf{0.828} & 0.028 & 0.181 & 0.838 \\
Mixed-FT (Ours) & 0.063 & 0.170 & 0.811 & \textbf{0.026} & \textbf{0.179} & \textbf{0.841} \\
\bottomrule
\end{tabular}
\end{table}

\begin{table}[ht]
\centering
\scriptsize
\caption{Structural fidelity on \textbf{ImageNet} via grayscale reprojection (see Table~\ref{tab:fidelity-coco-pixel} for the metric definition). Best in bold.}
\label{tab:fidelity-imagenet-pixel}
\setlength{\tabcolsep}{5pt}
\begin{tabular}{l ccc ccc}
\toprule
\multirow{2}{*}{Method} & \multicolumn{3}{c}{\textbf{Ortho}} & \multicolumn{3}{c}{\textbf{Pan}} \\
\cmidrule(lr){2-4} \cmidrule(lr){5-7}
& L1 $\downarrow$ & L2 $\downarrow$ & SSIM $\uparrow$ & L1 $\downarrow$ & L2 $\downarrow$ & SSIM $\uparrow$ \\
\midrule
Not-FT          & 0.126 & 0.270 & 0.373 & 0.128 & 0.271 & 0.370 \\
Normal-FT       & \textbf{0.050} & \textbf{0.133} & \textbf{0.857} & 0.026 & 0.136 & 0.875 \\
Mixed-FT (Ours) & 0.061 & 0.138 & 0.842 & \textbf{0.024} & \textbf{0.135} & \textbf{0.878} \\
\bottomrule
\end{tabular}
\end{table}

\begin{table}[ht]
\centering
\scriptsize
\caption{Structural fidelity on \textbf{Multi-Instance} via grayscale reprojection (see Table~\ref{tab:fidelity-coco-pixel} for the metric definition). Best in bold.}
\label{tab:fidelity-multi-instance-pixel}
\setlength{\tabcolsep}{5pt}
\begin{tabular}{l ccc ccc}
\toprule
\multirow{2}{*}{Method} & \multicolumn{3}{c}{\textbf{Ortho}} & \multicolumn{3}{c}{\textbf{Pan}} \\
\cmidrule(lr){2-4} \cmidrule(lr){5-7}
& L1 $\downarrow$ & L2 $\downarrow$ & SSIM $\uparrow$ & L1 $\downarrow$ & L2 $\downarrow$ & SSIM $\uparrow$ \\
\midrule
Not-FT          & 0.154 & 0.330 & 0.342 & 0.142 & 0.308 & 0.345 \\
Normal-FT       & \textbf{0.050} & \textbf{0.157} & \textbf{0.839} & 0.027 & 0.165 & 0.855 \\
Mixed-FT (Ours) & 0.060 & 0.161 & 0.826 & \textbf{0.025} & \textbf{0.163} & \textbf{0.857} \\
\bottomrule
\end{tabular}
\end{table}

\section{Conclusion}
We revisited image colorization through the lens of luminance mismatch. Most existing systems predict chroma while preserving an input-derived luminance channel, which works well when grayscale intensities align but can break under domain shifts such as historical orthochromatic imagery. To address this, we proposed a luminance-agnostic colorization framework that predicts full RGB outputs directly using a foundation image-editing model, rather than relying on luminance swapping.

Central to our approach is a mixed grayscale objective that trains the model under both standard panchromatic and simulated orthochromatic conditioning, allowing it to reinterpret brightness when the grayscale formation deviates from human-perceived luminance. Across COCO, ImageNet, and a multi-instance benchmark, we find that while our method remains competitive on standard grayscale inputs, it is substantially more robust on orthochromatic inputs, where prior methods degrade more strongly. Qualitative comparisons further indicate cleaner outputs with fewer local color artifacts.

Limitations: We adapt the foundation model using LoRA only; full fine-tuning with larger datasets, longer schedules, and more compute may further improve fidelity and robustness. Moreover, our orthochromatic simulation is a simplified approximation of historical film response, while real archival imagery exhibits diverse capture and aging effects (e.g., blue-biased spectral sensitivity, color fading, and chemical degradation) as well as material-dependent appearance changes. For instance, facial skin can appear harsher under red-insensitive imaging, plausibly due to reduced contributions from red-wavelength subsurface light transport. Future work includes incorporating more physically grounded spectral sensitivity models, training with a broader family of grayscale/color distortions, and extending the framework to jointly address additional restoration factors beyond grayscale formation.

\section{Acknowledgment}

The authors sincerely thank the conference reviewers for their valuable feedback. We also thank Krishn Vishwas Kher, Panshul Jindal, and Rajat Maheshwari for their valuable support. The first author, Swarnim Maheshwari, gratefully acknowledges support from the Reliance Foundation Fellowship. We thank Fujitsu for their support of our research by providing the necessary GPU compute.


%
%
\bibliographystyle{splncs04}
\bibliography{main}

@String(CVPR  = {IEEE Conf. Comput. Vis. Pattern Recog.})

@String(ICCV  = {Int. Conf. Comput. Vis.})

@String(ECCV  = {Eur. Conf. Comput. Vis.})

@String(ICLR  = {Int. Conf. Learn. Represent.})

@String(ICIP  = {IEEE Int. Conf. Image Process.})

@String(TOG   = {ACM Trans. Graph.})

@String(MM    = {ACM Int. Conf. Multimedia})

@String(NIPS  = {Adv. Neural Inform. Process. Syst.})

@String(WACV  = {IEEE Winter Conf. Appl. Comput. Vis.})

@String(CVPR  = {CVPR})

@String(ICCV  = {ICCV})

@String(ECCV  = {ECCV})

@String(ICLR  = {ICLR})

@String(ICIP  = {ICIP})

@String(TOG   = {ACM TOG})

@String(MM    = {ACM MM})

@String(NIPS  = {NeurIPS})

@String(WACV  = {WACV})

@InProceedings{lcoins,
  author = {Chang, Zheng and Weng, Shuchen and Zhang, Peixuan and Li, Yu and Li, Si and Shi, Boxin},
  title = {L-CoIns: Language-based Colorization with Instance Awareness},
  booktitle = cvpr,
  year = {2023}
}

@InProceedings{lcad,
  author = {Chang, Zheng and Weng, Shuchen and Zhang, Peixuan and Li, Yu and Li, Si and Shi, Boxin},
  title = {L-CAD: Language-based Colorization with Any-level Descriptions using Diffusion Priors},
  booktitle = nips,
  year={2023}
}

@inproceedings{coco-lc,
  title={COCO-LC: Colorfulness Controllable Language-based Colorization},
  author={Li, Yifan and Bai, Yuhang and Yang, Shuai and Liu, Jiaying},
  booktitle=mm,
  year={2024}
}

@article{unicolor,
  title={Unicolor: A unified framework for multi-modal colorization with transformer},
  author={Huang, Zhitong and Zhao, Nanxuan and Liao, Jing},
  journal=tog,
  year={2022},
}

@article{CtrlColor,
  title={Control Color: Multimodal Diffusion-based Interactive Image Colorization},
  author={Liang, Zhexin and Li, Zhaochen and Zhou, Shangchen and Li, Chongyi and Loy, Chen Change},
  journal={arXiv:2402.10855},
  year={2024}
}

@inproceedings{fid,
  title={Gans trained by a two time-scale update rule converge to a local nash equilibrium},
  author={Heusel, Martin and Ramsauer, Hubert and Unterthiner, Thomas and Nessler, Bernhard and Hochreiter, Sepp},
  booktitle=nips,
  year={2017}
}

@inproceedings{coco-stuff,
  title={Coco-stuff: Thing and stuff classes in context},
  author={Caesar, Holger and Uijlings, Jasper and Ferrari, Vittorio},
  booktitle={Proceedings of the IEEE conference on computer vision and pattern recognition},
  pages={1209--1218},
  year={2018}
}

@inproceedings{Diffusing-Colors,
author = {Zabari, Nir and Azulay, Aharon and Gorkor, Alexey and Halperin, Tavi and Fried, Ohad},
title = {Diffusing Colors: Image Colorization with Text Guided Diffusion},
year = {2023},
booktitle = {SIGGRAPH Asia 2023 Conference Papers},
articleno = {61},
numpages = {11},
}

@inproceedings{hasler2003measuring,
  title={Measuring colorfulness in natural images},
  author={Hasler, David and Suesstrunk, Sabine E},
  booktitle={Human vision and electronic imaging VIII},
  volume={5007},
  pages={87--95},
  year={2003},
  organization={SPIE}
}

@article{piggybacked,
  title={Improved diffusion-based image colorization via piggybacked models},
  author={Liu, Hanyuan and Xing, Jinbo and Xie, Minshan and Li, Chengze and Wong, Tien-Tsin},
  journal={arXiv preprint arXiv:2304.11105},
  year={2023}
}

@inproceedings{controlnet,
  title={Adding conditional control to text-to-image diffusion models},
  author={Zhang, Lvmin and Rao, Anyi and Agrawala, Maneesh},
  booktitle=cvpr,
  year={2023}
}

@inproceedings{cic,
  title={Colorful image colorization},
  author={Zhang, Richard and Isola, Phillip and Efros, Alexei A},
  booktitle=eccv,
  year={2016},
}

@misc{DeOldify,
  author = {Jason Antic},
  title = {DeOldify: A Deep Learning Based Project for Colorizing and Restoring Old Images (and Video!)},
  year = {2019},
  url={https://github.com/jantic/DeOldify},
  note = {Accessed: 2026-02-15},
}

@misc{MDL_GlassPlates,
  author        = {{Minnesota Digital Library}},
  title         = {Glass plate negatives collection},
  howpublished = {Minnesota Digital Library},
  year         = {2024},
  url          = {https://collection.mndigital.org/?f%5Bphysical_format_ssi%5D%5B%5D=Glass+plate+negatives},
  note         = {Accessed: 2026-02-15}
}

@misc{CIELab,
  title        = {Colorimetry---Part 4: CIE 1976 L*a*b* Colour Space},
  author       = {{Commission Internationale de l'{\'E}clairage (CIE)}},
  number       = {CIE S 014-4/E:2007},
  year         = {2007},
  institution  = {Commission Internationale de l'{\'E}clairage},
  url          = {https://cie.co.at/publications/colorimetry-part-4-cie-1976-lab-colour-space-1},
  urldate      = {2026-02-15},
  note         = {Standard CIE S 014-4/E:2007. Accessed: 2026-02-15}
}

@article{Luo-Rephotography-2021,
  author    = {Luo, Xuan and Zhang, Xuaner and Yoo, Paul and Martin-Brualla, Ricardo and Lawrence, Jason and Seitz, Steven M.},
  title     = {Time-Travel Rephotography},
  journal = {ACM Transactions on Graphics (Proceedings of ACM SIGGRAPH Asia 2021)},
  publisher = {ACM New York, NY, USA},
  volume = {40},
  number = {6},
  articleno = {213},
  doi = {https://doi.org/10.1145/3478513.3480485},
  year = {2021},
  month = {12}
}

@misc{oquab2023dinov2,
  title={DINOv2: Learning Robust Visual Features without Supervision},
  author={Oquab, Maxime and Darcet, Timothée and Moutakanni, Theo and Vo, Huy V. and Szafraniec, Marc and Khalidov, Vasil and Fernandez, Pierre and Haziza, Daniel and Massa, Francisco and El-Nouby, Alaaeldin and Howes, Russell and Huang, Po-Yao and Xu, Hu and Sharma, Vasu and Li, Shang-Wen and Galuba, Wojciech and Rabbat, Mike and Assran, Mido and Ballas, Nicolas and Synnaeve, Gabriel and Misra, Ishan and Jegou, Herve and Mairal, Julien and Labatut, Patrick and Joulin, Armand and Bojanowski, Piotr},
  journal={arXiv:2304.07193},
  year={2023}
}

@inproceedings{li2023blip2,
  title={Blip-2: Bootstrapping language-image pre-training with frozen image encoders and large language models},
  author={Li, Junnan and Li, Dongxu and Savarese, Silvio and Hoi, Steven},
  booktitle={International conference on machine learning},
  pages={19730--19742},
  year={2023},
  organization={PMLR}
}

@inproceedings{colorformer,
  title={ColorFormer: Image colorization via color memory assisted hybrid-attention transformer},
  author={Ji, Xiaozhong and Jiang, Boyuan and Luo, Donghao and Tao, Guangpin and Chu, Wenqing and Xie, Zhifeng and Wang, Chengjie and Tai, Ying},
  booktitle=eccv,
  year={2022},
}

@inproceedings{HistoryNet,
  title = {Focusing on Persons: Colorizing Old Images Learning from Modern Historical Movies},
  author = {Jin, Xin and Li, Zhonglan and Liu, Ke and Zou, Dongqing and Li, Xiaodong and Zhu, Xingfan and Zhou, Ziyin and Sun, Qilong and Liu, Qingyu},
  booktitle = mm,
  year = {2021},
}

@misc{pyiqa,
  title={{IQA-PyTorch}: PyTorch Toolbox for Image Quality Assessment},
  author={Chaofeng Chen and Jiadi Mo},
  year={2022},
  howpublished = "[Online]. Available: \url{https://github.com/chaofengc/IQA-PyTorch}",
  note = {Accessed: 2026-02-15}
}

@misc{kaggle_imagenet,
  title={ImageNet Object Localization Challenge},
  author={{Kaggle}},
  year={2020},
  url = "https://www.kaggle.com/competitions/imagenet-object-localization-challenge/data",
  note = {Accessed: 2026-02-15}
}

@inproceedings{tovivid,
  title={Towards vivid and diverse image colorization with generative color prior},
  author={Wu, Yanze and Wang, Xintao and Li, Yu and Zhang, Honglun and Zhao, Xun and Shan, Ying},
  booktitle=cvpr,
  year={2021}
}

@misc{kodak_mpt_glossary_ortho,
  title        = {Glossary of Motion Picture Terms},
  author       = {{Kodak}},
  year         = {n.d.},
  url          = {https://www.kodak.com/en/motion/page/glossary-of-motion-picture-terms/},
  urldate      = {2026-02-15},
  note         = {Entry: ``Orthochromatic (Ortho) Film'' (sensitive to only blue and green light). Accessed: 2026-02-15.}
}

@misc{kodak_filmessentials_characteristics_06,
  title        = {Basic Sensitometry and Characteristics of Film (Film Essentials, Module 6)},
  author       = {{Kodak}},
  year         = {n.d.},
  url          = {https://www.kodak.com/uploadedfiles/motion/US_plugins_acrobat_en_motion_newsletters_filmEss_06_Characteristics_of_Film.pdf},
  urldate      = {2026-02-15},
  note         = {States orthochromatic films are sensitive mainly to the blue-green portions of the visible spectrum. Accessed: 2026-02-15.}
}

@misc{britannica_orthochromatic_film,
  title        = {Orthochromatic Film},
  author       = {{Encyclopaedia Britannica}},
  year         = {n.d.},
  url          = {https://www.britannica.com/technology/orthochromatic-film},
  urldate      = {2026-02-15},
  note         = {Describes orthochromatic films as sensitive to violet/blue/green/yellow but not to red. Accessed: 2026-02-15.}
}

@misc{filmcolors_orthochromatic_stock_1873,
  title        = {Orthochromatic Stock},
  author       = {{Film Colors}},
  year         = {n.d.},
  url          = {https://filmcolors.org/timeline-entry/1346/},
  urldate      = {2026-02-15},
  note         = {Historical note: sensitivity extended to record green as well as blue (Vogel, 1873). Accessed: 2026-02-15.}
}

@inproceedings{
coltran,
title={Colorization Transformer},
author={Manoj Kumar and Dirk Weissenborn and Nal Kalchbrenner},
booktitle=iclr,
year={2021},
}

@article{DISCO,
author   = {Menghan Xia and Wenbo Hu and Tien Tsin Wong and Jue Wang},
title    = {Disentangled Image Colorization via Global Anchors},
journal  = tog,
year = {2022}
}

@InProceedings{ct2,
  author = {Weng, Shuchen and Sun, Jimeng and Li, Yu and Li, Si and Shi, Boxin},
  title = {CT2: Colorization Transformer via Color Tokens},
  booktitle = eccv,
  year = {2022}
}

@inproceedings{ddcolor,
  title={DDColor: Towards Photo-Realistic Image Colorization via Dual Decoders},
  author={Kang, Xiaoyang and Yang, Tao and Ouyang, Wenqi and Ren, Peiran and Li, Lingzhi and Xie, Xuansong},
  booktitle=iccv,
  year={2023}
}

@inproceedings{instColor,
  author = {Su, Jheng-Wei and Chu, Hung-Kuo and Huang, Jia-Bin},
  title = {Instance-aware Image Colorization},
  booktitle = cvpr,
  year = {2020}
}

@InProceedings{ChromaGAN,
  author = {Vitoria, Patricia and Raad, Lara and Ballester, Coloma},
  title = {ChromaGAN: Adversarial Picture Colorization with Semantic Class Distribution},
  booktitle = wacv,
  month = {March},
  year = {2020}
}

@inproceedings{bigcolor,
  title={BigColor: Colorization using a generative color prior for natural images},
  author={Kim, Geonung and Kang, Kyoungkook and Kim, Seongtae and Lee, Hwayoon and Kim, Sehoon and Kim, Jonghyun and Baek, Seung-Hwan and Cho, Sunghyun},
  booktitle=eccv,
  year={2022},
}

@article{biggan,
  title={Large scale GAN training for high fidelity natural image synthesis},
  author={Brock, Andrew and Donahue, Jeff and Simonyan, Karen},
  journal={arXiv preprint arXiv:1809.11096},
  year={2018}
}

@article{MultiColor,
  title={MultiColor: Image Colorization by Learning from Multiple Color Spaces},
  author={Du, Xiangcheng and Zhou, Zhao and Wang, Yanlong and Wang, Zhuoyao and Zheng, Yingbin and Jin, Cheng},
  journal={arXiv preprint arXiv:2408.04172},
  year={2024}
}

@article{imagenet,
  title={Imagenet large scale visual recognition challenge},
  author={Russakovsky, Olga and Deng, Jia and Su, Hao and Krause, Jonathan and Satheesh, Sanjeev and Ma, Sean and Huang, Zhiheng and Karpathy, Andrej and Khosla, Aditya and Bernstein, Michael and others},
  journal={International journal of computer vision},
  volume={115},
  pages={211--252},
  year={2015},
  publisher={Springer}
}

@article{sfidding2020continuous,
  title={Continuous conditional generative adversarial networks for image generation: Novel losses and label input mechanisms},
  author={Ding, Xin and Wang, Yongwei and Xu, Zuheng and Welch, William J and Wang, Z Jane},
  journal={arXiv preprint arXiv:2011.07466},
  year={2020}
}

@inproceedings{zerman2019colornet,
  title={Colornet-estimating colorfulness in natural images},
  author={Zerman, Emin and Rana, Aakanksha and Smolic, Aljosa},
  booktitle={2019 IEEE International Conference on Image Processing (ICIP)},
  pages={3791--3795},
  year={2019},
  organization={IEEE}
}

@inproceedings{bozic2024versatile,
  title={Versatile vision foundation model for image and video colorization},
  author={Bozic, Vukasin and Djelouah, Abdelaziz and Zhang, Yang and Timofte, Radu and Gross, Markus and Schroers, Christopher},
  booktitle={ACM SIGGRAPH 2024 Conference Papers},
  pages={1--11},
  year={2024}
}

@misc{flux-2-2025,
    author={Black Forest Labs},
    title={{FLUX.2: Frontier Visual Intelligence}},
    year={2025},
    howpublished={\url{https://bfl.ai/blog/flux-2}},
    note={Accessed: 2026-02-15},
}

@misc{labs2025flux1kontextflowmatching,
      title={FLUX.1 Kontext: Flow Matching for In-Context Image Generation and Editing in Latent Space},
      author={Black Forest Labs and Stephen Batifol and Andreas Blattmann and Frederic Boesel and Saksham Consul and Cyril Diagne and Tim Dockhorn and Jack English and Zion English and Patrick Esser and Sumith Kulal and Kyle Lacey and Yam Levi and Cheng Li and Dominik Lorenz and Jonas Müller and Dustin Podell and Robin Rombach and Harry Saini and Axel Sauer and Luke Smith},
      year={2025},
      eprint={2506.15742},
      archivePrefix={arXiv},
      primaryClass={cs.GR},
      url={https://arxiv.org/abs/2506.15742},
}

@misc{wu2025qwenimagetechnicalreport,
      title={Qwen-Image Technical Report}, 
      author={Chenfei Wu and Jiahao Li and Jingren Zhou and Junyang Lin and Kaiyuan Gao and Kun Yan and Sheng-ming Yin and Shuai Bai and Xiao Xu and Yilei Chen and Yuxiang Chen and Zecheng Tang and Zekai Zhang and Zhengyi Wang and An Yang and Bowen Yu and Chen Cheng and Dayiheng Liu and Deqing Li and Hang Zhang and Hao Meng and Hu Wei and Jingyuan Ni and Kai Chen and Kuan Cao and Liang Peng and Lin Qu and Minggang Wu and Peng Wang and Shuting Yu and Tingkun Wen and Wensen Feng and Xiaoxiao Xu and Yi Wang and Yichang Zhang and Yongqiang Zhu and Yujia Wu and Yuxuan Cai and Zenan Liu},
      year={2025},
      eprint={2508.02324},
      archivePrefix={arXiv},
      primaryClass={cs.CV},
      url={https://arxiv.org/abs/2508.02324}, 
}
\end{document}